\documentclass[10pt,journal,compsoc]{IEEEtran}
\usepackage[nocompress]{cite}

\usepackage[margin=0pt,font=small,labelfont=bf,labelsep=endash,tableposition=top]{caption}

\usepackage{amsmath}
\usepackage{array}

\usepackage[caption=false,font=footnotesize,labelfont=sf,textfont=sf]{subfig}
\usepackage{fixltx2e}

\usepackage{url}
\usepackage{graphicx}
\usepackage{amsmath}
\usepackage{amssymb}

\usepackage{epsfig}
\usepackage{epstopdf}
\usepackage{makecell,multirow,diagbox}
\usepackage{color}
\usepackage{soul}
\usepackage{url}
\usepackage{amsmath}
\usepackage{array}

\usepackage[utf8]{inputenc}
\usepackage{xcolor}
\usepackage{hyperref}
\usepackage{helvet}
\usepackage{courier}
\usepackage{bm}
\usepackage{bbm}
\usepackage{wrapfig}
\usepackage{picinpar}
\usepackage{arydshln}

\usepackage[vlined,ruled,linesnumbered]{algorithm2e}
\usepackage{cite}
\usepackage{balance}
\usepackage{amssymb}

\usepackage{mathtools,xspace}

\def\etal{{\it et al.}\xspace}

\hypersetup{
    colorlinks=true,
    breaklinks=true,
    urlcolor= blue,
    linkcolor= red,
    bookmarksopen=false,
    filecolor=black,
    citecolor=blue,
    linkbordercolor=blue
}
\AtBeginDocument{\hypersetup{pdfborder={0 0 1}}}%

\begin{document}

\title{ABCNet v2: Adaptive Bezier-Curve Network for Real-time End-to-end Text Spotting}

\author{
Yuliang Liu$ ^{\ddag\dag}$%
, Chunhua Shen$ ^{\dag*}$,
Lianwen Jin$ ^{\ddag*}$,
Tong He$ ^\dag$,
Peng Chen$ ^\dag$,
Chongyu Liu$ ^\ddag$,
Hao Chen$ ^\dag$
\thanks{
$ ^\ddag$ South China University of Technology;
$ ^\dag$ The University of Adelaide, Australia.
}
\thanks{$^*$ Corresponding authors.}
}

\IEEEtitleabstractindextext{%
\begin{abstract}

End-to-end text-spotting, which aims to integrate detection and recognition in a unified framework, has attracted increasing attention due to its simplicity of the two complimentary tasks. It remains an open problem especially when processing arbitrarily-shaped text instances. Previous methods can be roughly categorized into two groups: character-based and segmentation-based, which often require character-level annotations and/or complex post-processing due to the unstructured output. Here, we tackle end-to-end text spotting by presenting Adaptive Bezier Curve Network v2 (ABCNet v2). Our main contributions are four-fold: 1) For the first time, we adaptively fit arbitrarily-shaped text by a parameterized Bezier curve, which, compared with segmentation-based methods, can not only provide structured output but also controllable representation. 2) We design a novel BezierAlign layer for extracting accurate convolution features of a text instance of arbitrary shapes, significantly improving the precision of recognition over previous methods. 3) Different from previous methods, which often suffer from complex post-processing and sensitive hyper-parameters, our ABCNet v2 maintains a simple pipeline with the only post-processing non-maximum suppression (NMS). 4) As the performance of text recognition closely depends on feature alignment, ABCNet v2 further adopts a simple yet effective coordinate convolution to encode the position of the convolutional filters, which leads to a considerable improvement with negligible computation overhead. Comprehensive experiments conducted on various bilingual (English and Chinese) benchmark datasets demonstrate that ABCNet v2 can achieve state-of-the-art performance while maintaining very high efficiency. More importantly, as there is little work on  quantization of text spotting models, we quantize our models to improve the inference time of the proposed ABCNet v2. This can be valuable for real-time applications. Code and model are available at:   \url{https://git.io/AdelaiDet}

\end{abstract}

}

\maketitle

\IEEEdisplaynontitleabstractindextext
\IEEEpeerreviewmaketitle

\IEEEraisesectionheading{\section{Introduction}\label{sec:introduction}}
\IEEEPARstart{T}{ext} spotting in the natural environment has drawn increasing research attention in the community of computer vision and image understanding, which aims to detect and recognize text instances in unconstrained conditions. Text information recovered has proven valuable for image retrieval, automatic organization of photos and visual assistance, to name a few. To date, it remains challenging for the following reasons:
1) text instances often exhibit diverse patterns in shape, color, font, and language. These inevitable variations in the data often require heuristic settings for achieving satisfactory performance;
2) real-time applications require the algorithm to achieve a better trade-off between the efficiency and effectiveness. Although the emergence of deep learning has significantly improved the performance of the task of scene text spotting, current methods still exist a considerable gap for solving generic real-world applications.

Previous approaches often involve two independent modules for scene text spotting: text detection and recognition, which are implemented sequentially. Many of them only address one task and directly borrow top-performing modules from the other.
Such a simplified approach is unlikely to exploit the full potential of deep convolution networks, as two tasks are isolated without shareable features.

\begin{figure}[t!]
  \begin{minipage}[c]{0.49\linewidth}
    \centering
    \centerline{\includegraphics[width = 4.1cm, height = 3.5cm]{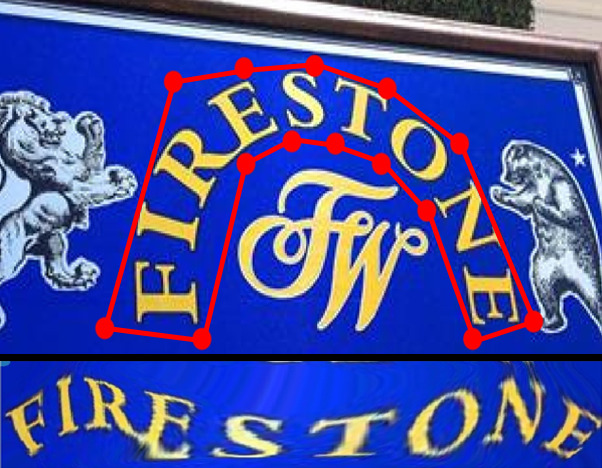}}
    \centerline{\small{(a) TPS-based alignment. }}
  \end{minipage}
  \begin{minipage}[c]{0.49\linewidth}
    \centering
    \centerline{\includegraphics[width = 4.1cm, height = 3.5cm]{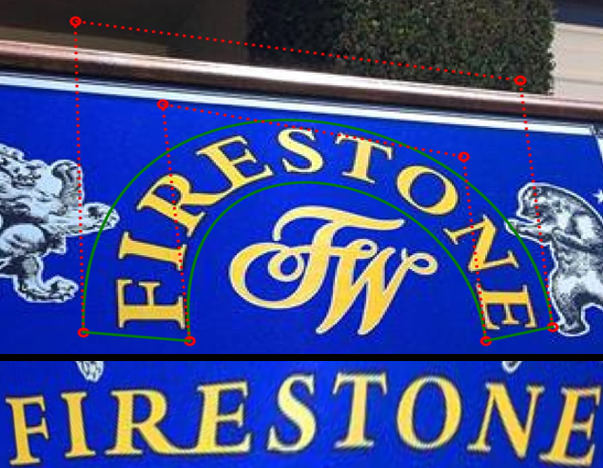}}
    \centerline{\small{(b) BezierAlign-ed results. }}
  \end{minipage}
  \caption{
  \textbf{Comparison of the warping results}. In Figure (a), we follow previous methods by using TPS \cite{bookstein1989principal} and STN \cite{jaderberg2015spatial} to warp the curved text region into a rectangular shape. In Figure (b), we use generated Bezier curves  and the proposed BezierAlign to warp the results, leading to improved accuracy. }\label{fig:Bezier_gen_comp}
\end{figure}

Recently, end-to-end scene text spotting methods  \cite{li2017towards,he2018end,liu2018fots,lyu2018mask,qin2019towards,xing2019convolutional,feng2019textdragon,liao2019mask,li2019towards}, which directly build the mapping between the input image and sets of words transcripts in a unified framework, is drawing increasing attention.
Compared to the models in which the detection and recognition are two separate modules, the advantages of designing an end-to-end framework are as follows.
Firstly, word recognition can significantly improve the accuracy of detection. One of the most notable characteristics for text is the
attribute of being sequences. However, false positives exhibiting the appearance of sequences exist in unconditioned environments, such as blocks, buildings, and railings. To empower the network to have discriminative capability to distinguish different patterns, some
approaches \cite{li2017towards,he2018end,liu2018fots} propose to share the features between the two tasks and train the network in an end-to-end manner.
Moreover, due to the shared features, end-to-end frameworks often show superiority in the inference speed, which is more suitable for real-time applications.
Finally, current standalone recognition models usually adopt perfectly cropped text images or heuristic synthetic images for training. An end-to-end module can force the recognition module to accustom to the detection outputs, and thus the results can be more robust \cite{liao2019mask}.

Existing approaches for end-to-end text-spotting can be roughly categorized into two groups: character-based and segmentation-based. Character-based methods first detect and recognize individual characters and then output words by applying an extra grouping module. Although effective, laborious character-level annotations are required. Besides,
often a few predefined hyper-parameters are necessary for the grouping algorithm, showing limited robustness and generalization capability. Another line of the research is segmentation based, where text instances are represented by unstructured contours, making it difficult for the
subsequent recognition step. For example, the work in \cite{boundary2020} relies on an TPS \cite{bookstein1989principal} or an STN \cite{jaderberg2015spatial} step to warp the original ground truths into rectangular shape.
Note that, the characters can be significantly distorted, as shown in Figure \ref{fig:Bezier_gen_comp}.
Besides, compared with detection, text recognition requires a significantly large amount of training data, resulting in optimization difficulties in a unified framework.

To address these limitations, we propose the Adaptive Bezier Curve Network v2 (ABCNet v2), an end-to-end trainable framework, for real-time arbitrarily shaped scene text spotting. ABCNet v2 enables arbitrarily shaped scene text detection with simple yet effective Bezier curve adaptation, which introduces negligible computation overhead compared with standard rectangle bounding box detection. In addition, we design a novel feature alignment layer, termed BezierAlign, to precisely calculate convolutional features of text instances in curved shapes, and thus high recognition accuracy can be achieved.
\textit{For the first time, we successfully adopt the parameter space (Bezier curves) for multi-oriented or curved text spotting, enabling a very concise and efficient pipeline.}

Inspired by recent work in \cite{tan2020efficientdet,liu2018intriguing,wang2020solov2},
we improve ABCNet in our conference version  \cite{liu2020abcnet} in four aspects: the feature extractor, detection branch, recognition branch, and end-to-end training. Due to the inevitable variation of scales, the proposed ABCNet v2 incorporates iterative bidirectional features to achieve a better accuracy and efficiency trade-off.
In addition, based on our observations, feature alignment in the detection branch is essential for the subsequent text recognition. To this end, we adopt a coordinate encoding approach with negligible computation overhead to explicitly encode the position in
the convolutional filters, leading to considerable improvement in accuracy.
For the recognition branch, we integrate a character attention module, which can recursively predict the characters of each word without using character-level annotations. To enable effective end-to-end training, we further propose an Adaptive End-to-End Training (AET) strategy to match detection for end-to-end training.
This can force the recognition branch more robust to the detection behaviors.

Thus, the proposed ABCNet v2 enjoys several advantages over previous state-of-the-art methods, which are summarized as follows:
\begin{itemize}
    \item

For the first time, we introduce a new, concise parametric representation of curved scene text using Bezier curves. It introduces negligible computation overhead compared with the standard bounding box representation.

    \item

We propose a new feature alignment method, \textit{a.k.a.}\ BezierAlign,
and thus the recognition branch can be seamlessly connected to the overall structure. By sharing backbone features, the recognition branch can be designed with a light-weight structure for efficient inference.

    \item

The detection model of ABCNet v2 is more general for processing
multi-scale text instances by considering bidirectional multi-scale pyramid global textual features.

    \item

To our knowledge, our method is the first framework that can simultaneously detect and recognize horizontal, multi-oriented, and arbitrarily shaped text in a single-shot manner, while keeping a real-time inference speed.

    \item

To further speed up inference, we also exploit the technique of
model quantization, showing that ABCNet v2 can reach a much
faster inference speed with only marginal accuracy reduction.

    \item

Comprehensive experiments on various benchmarks demonstrate the state-of-the-art text spotting performance of the proposed ABCNet v2 in
terms of accuracy and speed.

\end{itemize}

\section{Related Work}
\label{sec:related-works}
Scene text spotting requires detecting and recognizing text simultaneously, instead of concerning only one task. In the early days, scene text spotting methods are usually simply connected by the independent detection and recognition models. Two models are separately optimized with different architectures. Recently, end-to-end methods (\S \ref{subsec:re_e2e})  have significantly advanced the performance of text spotting by integrating the detection and recognition into one unified network.

\subsection{Separate Scene Text Spotting}
In this section, we briefly review the literature, focusing  on either detection or recognition.

\subsubsection{Scene Text Detection}
The development trend of text detection can be observed through the detection flexibility. From focused horizontal scene text detection represented by horizontal rectangular detection bounding boxes, to multi-oriented scene text detection represented by rotated rectangular or quadrilateral bounding boxes, and to arbitrarily shaped scene text detection represented by instance segmentation masks or polygons.

The early horizontal rectangular based methods can be dated back to Lucas    \etal     \cite{lucas2003icdar}, in which the pioneering  horizontal ICDAR'03 benchmark is constructed. ICDAR'03 and its successive datasets (ICDAR'11 \cite{shahab2011icdar} and ICDAR'13 \cite{Karatzas2013ICDAR}) have attracted considerable research efforts \cite{epshtein2010detecting,huang2013text,huang2014robust,liang2015multi,liao2017textboxes,tian2016detecting} in studies on horizontal scene text detection.

Before 2010, most of the methods merely focus on regular horizontal scene text, which is limited to generalize to the real applications, where multi-oriented scene text is ubiquitous. To this end, Yao    \etal     \cite{yao2012detecting} put forward a practical detection system as well as a multi-oriented benchmark (MSRA-TD500) for multi-oriented scene text detection. Both the method and the dataset use rotated rectangular bounding boxes to detect and annotate the multi-oriented text instances. Besides MSRA-TD500, the emergence of other multi-oriented datasets including NEOCR \cite{nagy2011neocr}, and USTB-SV1K \cite{yin2015multi} further facilitate numerous rotated rectangle based methods \cite{shivakumara2010laplacian,yao2012detecting,yin2015multi,yin2013robust,zhang2016multi,li2017towards}. Since 2015, ICDAR'15 \cite{karatzas2015icdar} starts to use four points based quadrilateral annotations for each text instance, which facilitates numerous methods that successfully demonstrate the superiority of the tighter and more flexible quadrilateral detection methods. The SegLink method \cite{shi2017detecting} predicts text regions by oriented segments and learns connecting links to recombine the results. DMPNet \cite{liu2017deep} observes that rotated rectangles may still contain
unnecessary background noises, imperfect matching, or unnecessary overlap, and thus it proposes the use of quadrilateral bounding boxes to detect text with auxiliary, predefined quadrilateral sliding windows.
EAST \cite{zhou2017east} employs a dense prediction structure for directly predicting quadrilateral bounding boxes. WordSup \cite{hu2017wordsup} proposes an iterative strategy to generate characters region automatically, which shows robustness on complicated scenes. The successful attempt of the ICDAR 2015 motivates numerous quadrilateral based datasets, such as RCTW'17 \cite{shi2017icdar2017}, MLT \cite{nayef2019icdar2019}, and ReCTS \cite{zhang2019icdar}.

Recently, the research focus has shifted from multi-oriented scene text detection to arbitrarily shaped text detection. The arbitrary shape is mainly presented by curved text in the wild, which can also be very common in our real world, \textit{e.g.}, columnar objects (bottles and stone piles), spherical objects, plicate planes (clothes, streamer, and receipts), coins, logos, seal, signboard and so on. The first curved text dataset CUTE80 \cite{risnumawan2014robust} is constructed in 2014.
But this dataset is mainly used for scene text recognition, as it contains only 80 clean images with relative few text instances. For detection
on arbitrarily shaped scene text, two recent benchmarks---Total-Text \cite{ch2019total} and SCUT-CTW1500 \cite{liu2019curved}---have been proposed to advance many influential works \cite{liao2020real, baek2019character, wang2019efficient,wang2019shape,long2018textsnake,tang2019seglink++,zhang2019look,wang2020contournet,xue2019msr}.
TextSnake \cite{long2018textsnake} designs an FCN to predict the geometry attributes of text instances and then groups them into the final output. CRAFT \cite{baek2019character} predicts the character regions of the text and the affinity between the adjacent ones. SegLink++ \cite{tang2019seglink++} proffers an instance-aware component grouping framework for dense and arbitrary shaped text detection. PSENet \cite{wang2019shape} proposes to learn the text kernel, then expands them to cover the whole text instances. PAN \cite{wang2019efficient} based on PSENet \cite{wang2019shape}, adopts a learnable post-processing method by predicting similarity vectors of pixels.
Wang    \etal     \cite{wang2019arbitrary} propose to learn the adaptive text region representation for the detection of text in arbitrary shape.
DRRN \cite{zhang2020deep} proposes to first detect text components, and then groups them together through a graph network. ContourNet \cite{wang2020contournet} adopts an adaptive-RPN and an extra contour prediction branch to improve the precision.

\subsubsection{Scene Text Recognition}
Scene text recognition aims at recognizing text through a cropped text image. Many previous methods \cite{neumann2012real,yao2014strokelets}, following the bottom-up approach, first segment character regions through sliding windows and classify each character, then group them into a word for taking the dependence with its neighbors into consideration. They achieve good performance in scene text recognition, but are limited to costly character-level annotations for character detection.
Without large training datasets, the models' in this category typically cannot generalize well.
The works proposed by Su and Lu \cite{su2014accurate,su2017accurate} present a scene text recognition system by using HOG feature and Recurrent Neural Network (RNN), which are one of pioneer works that successfully introduce RNN for scene text recognition.
Later, CNN-based methods with recurrent neural network are proposed to perform in a top-down manner, which can end-to-end predict a text sequence without any character detection. Shi \etal     \cite{shi2017end} apply Connectionist Temporal Classification (CTC) \cite{graves2006connectionist} upon a network integrated CNNs with RNNs, termed CRNN. Guided by the CTC loss, CRNN-based model can effectively transcribe the image content. Besides CTC, the attention mechanism \cite{bahdanau2014neural} is also employed for text recognition.

The above methods are mainly applied to regular text recognition and
not sufficiently robust for irregular one. In recent years, approaches for arbitrary-shape  text recognition become dominant, which can be categorized into rectification-based methods and rectification-free methods. For the former, STN \cite{jaderberg2015spatial} and Thin-Plate-Spline (TPS) \cite{warps1989thin} are two widely used methods for text rectification. Shi    \etal     \cite{shi2016robust} are the first to introduce STN and the attention-based decoder for predicting the text sequence. The work in \cite{shi2018aster} achieves better performance using iterative text rectification. Besides, Luo    \etal     \cite{luo2019moran} propose MORAN, which rectifies  the text through regressing the offsets for location shift. Liu    \etal     \cite{liu2018char} propose a Character-Aware Neural Network (Char-Net), which detects characters first and then separately transforms them into horizontal one.
ESIR \cite{zhan2019esir}  presents an iterative rectification pipeline that can turn the position of text from perspective distortion to regular format, and thus an effective end-to-end scene text recognition system can be built.
Litman    \etal     \cite{Ron2020Scatter} first apply TPS on input images and then stack several selective attention decoder for both visual and contextual features.

In  the category of rectification-free methods, Cheng    \etal     \cite{cheng2018aon} propose an arbitrary orientation network (AON) to extract features in four directions and the character position clues. Li    \etal     \cite{li2019show} apply the 2D-attention mechanism to capture irregular text features and achieved impressive results.
To tackle the attention drift issue, Yue    \etal     \cite{robustscanner} design a novel position enhancement branch in the recognition model.
In addition, some rectification-free methods are based on semantic segmentation. Liao    \etal     \cite{liao2019scene} and Wan    \etal     \cite{wan2019textscanner} both propose to segment characters and classify their categories through visual features.

\subsection{End-to-End Scene Text Spotting}
\label{subsec:re_e2e}
\subsubsection{Regular End-to-end Scene Text Spotting}
Li    \etal    \ \cite{li2017towards} may be the first to propose an end-to-end trainable scene text spotting method. The method successfully uses an RoI Pooling \cite{ren2015faster} to join detection and recognition features via a two-stage framework.
It is designed to process horizontal and focused text.
Its improved version \cite{li2019towards} significantly improves the performance.
Busta    \etal     \cite{busta2017deep} also propose an end-to-end deep text spotter.
He    \etal     \cite{he2018end} and Liu    \etal    \ \cite{liu2018fots} adopt an anchor-free mechanism to improve both the training and inference speed. They use a similar sampling strategy, \textit{i.e.}, Text-Align-Sampling and RoI-Rotate, respectively, to enable extracting features from quadrilateral detection results.
Note that both of these two methods are not capable of spotting arbitrarily-shaped scene text.

\subsubsection{Arbitrarily-shaped End-to-end Scene Text Spotting}
To detect arbitrarily-shaped scene text, Liao    \etal     \cite{lyu2018mask} propose a Mask TextSpotter which subtly refines Mask R-CNN and uses character-level supervision to simultaneously detect and recognize characters and instance masks. The method significantly improves the performance of spotting arbitrarily-shaped scene text.
Its improved version \cite{liao2019mask} significantly alleviates  the reliance on the character-level annotations.
Sun    \etal     \cite{sun2018textnet} propose the TextNet which produces quadrilateral detection bounding boxes in advance, and then use a region proposal network to feed the detection features for recognition.

\begin{figure*}[!t]
   \centering
   \centerline{\includegraphics[width=17.6cm,
   ]{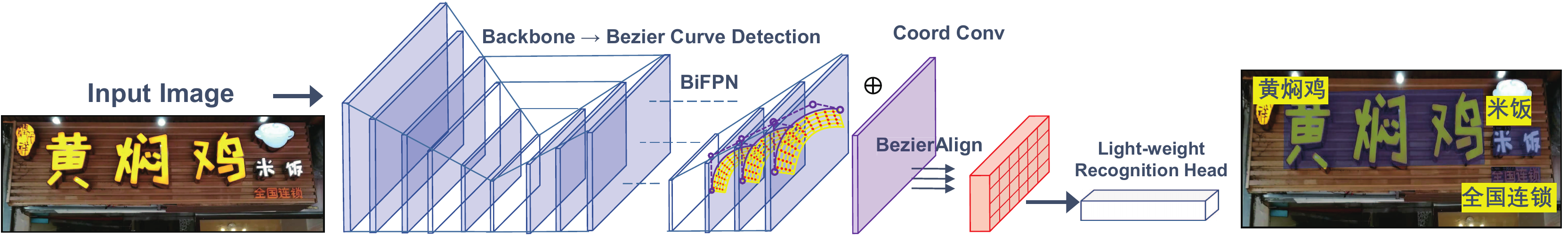}}
   \caption{
   \textbf{The framework of the proposed ABCNet v2}. We use cubic Bezier curves and BezierAlign to extract multi-scale curved sequence features using the Bezier curve detection results. We concatenate coordinate channels to encode the position coordinates in FPN output features before sending to BezierAlign. The overall framework is end-to-end trainable with high efficiency. Here purple dots represent the control points of the cubic Bezier curve. }\label{fig:pipeline}
\end{figure*}

Recently, Qin    \etal      \cite{qin2019towards} propose to use a RoI Masking to focus on the arbitrarily-shaped text region.
Note that extra computation is needed to fit polygons.
The work in \cite{xing2019convolutional} propose an arbitrarily-shaped scene text spotting method, termed CharNet, requiring character-level
training data and TextField \cite{xu2019textfield} to group recognition results.
Authors of \cite{feng2019textdragon} propose a novel sampling method, RoISlide, which uses fused features from the predicting segments of the text instances, and thus it is robust to long arbitrarily-shaped text.

Wang    \etal     \cite{boundary2020} first detect the boundary points of text in arbitrary shapes, through TPS to rectify the detected text and then fed it into the recognition branch.
Liao    \etal     \cite{liao2020masktext} propose a Segmentation Proposal Network (SPN) to accurate extract the text regions, and it follows \cite{liao2019mask} to attain the final results.

\section{Our Method}
\label{sec:method}
An intuitive pipeline of our method is shown in Figure \ref{fig:pipeline}. Inspired by \cite{zhong2019anchor,tian2019fcos,he2017deep}, we adopt a single-shot, anchor-free convolutional neural network as the detection framework. Removal of anchor boxes significantly simplifies the detection for our task. Here the detection is densely predicted on the output feature maps of the detection head, which is constructed by 4 stacked convolution layers with stride of 1, padding of 1, and 3$\times$3 kernels. Next, we present the key components of the proposed ABCNet v2 in six components:
1) Bezier curve detection; 2) the coordinate convolution module; 3) BezierAlign; 4) the light-weight attention recognition module; 5) the adaptive end-to-end training strategy; and 6) text spotting quantization.

\subsection{Bezier Curve Detection}
\label{subsec:Bezier_det}
Compared to segmentation-based methods \cite{wang2019shape,xu2019textfield,baek2019character,tian2019learning,
zhang2019look,long2018textsnake}, regression-based methods are more suitable for arbitrarily-shaped text detection, as demonstrated in \cite{liu2019curved,wang2019arbitrary}.
A drawback of these methods is the complicated pipeline, and they often require complex post-processing steps to obtain the final results.

\begin{figure}[!b]
\centering
\centerline{\includegraphics[width=0.5\textwidth%
]
{
        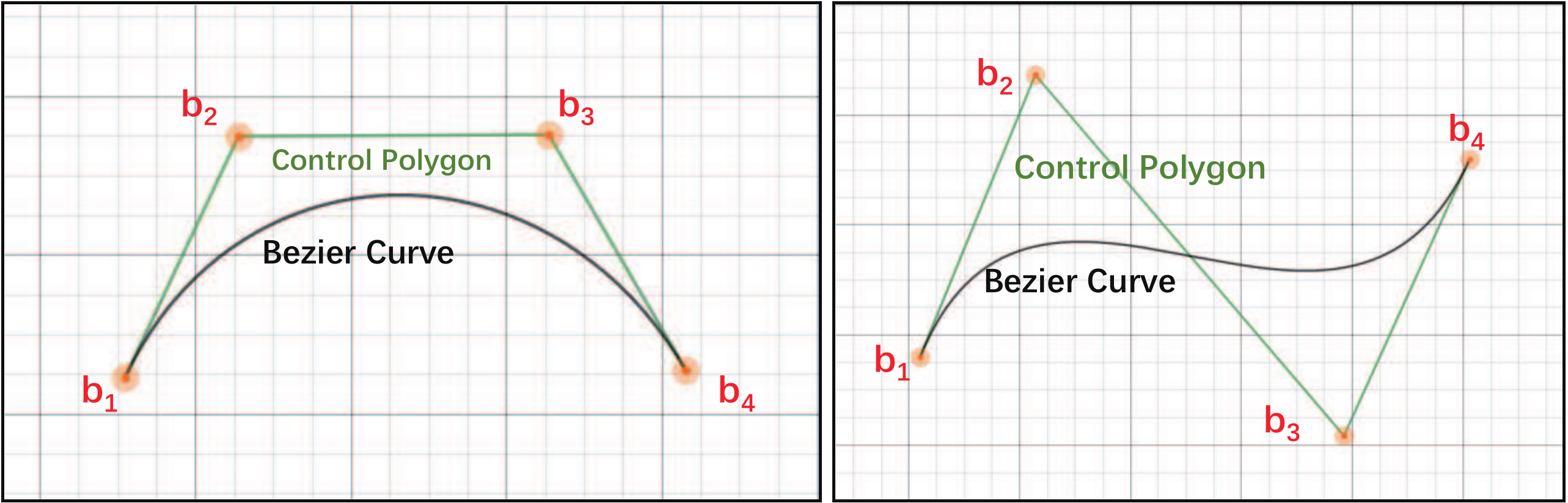}
}
   \caption
   {
    \textbf{Cubic Bezier curves.}
    $ b_{i} $ represents the control points. The green lines form a control polygon,  and the black curve is the cubic Bezier curve.
    Note that with only two end-points $ b_1 $ and $ b_4 $,
    the Bezier curve degenerates to a straight line.
}
    \label{fig:Bezier_ill}
\end{figure}

To simplify the detection for arbitrarily-shaped scene text instances, we propose to fit a Bezier curve by regressing several key points.
The Bezier curve represents a parametric curve  $c(t)$ that uses the Bernstein Polynomials as its basis. The definition is shown in Equation \eqref{eq:Bezier_def}:
\begin{equation}\label{eq:Bezier_def}
   c(t) = \sum_{i=0}^{n}b_{i}B_{i,n}(t), \;\;\;
   0\leq t\leq 1,
 \end{equation}
where, $n$ represents the degree, $b_{i}$ represents the $i$-$th$ control points, and $B_{i,n}(t)$ represents the Bernstein basis polynomials, as shown in Equation \eqref{eq:Bernstein}:
\begin{equation}\label{eq:Bernstein}
   B_{i,n}(t) = \binom{n}{i}t^{i}(1-t)^{n-i}, \;\;\;  i=0,...,n,
 \end{equation}
where $\binom{n}{i}$ is the binomial coefficient. To fit arbitrary shapes of the text with Bezier curves, we examine the arbitrarily-shaped scene text from the existing datasets and we empirically show that a cubic Bezier curve (\textit{i.e.}, $n = 3$) is sufficient to fit different
formats of curved scene text, especially on dataset with word-level annotation. A higher-order may work better on text-line level datasets, where multiple waves may be presented in one instance. We provide comparisons in terms of the order of the Bezier curves in the experiment section. An illustration of cubic Bezier curves is shown in Figure \ref{fig:Bezier_ill}.

Based on the cubic Bezier curve, we can formulate the arbitrarily-shaped scene text detection into a regression problem similar to bounding box regression, but with eight control points in total. Note that a straight text that has four control points (four vertexes) is a typical case of arbitrarily-shaped scene text. For consistency, we interpolate additional two control points in the tripartite points of each long side.

To learn the coordinates of the control points, we first generate the Bezier curve annotations described in \S\ref{subsubsec:Bezier_gen} and follow a similar regression method as in \cite{liu2017deep} to regress the targets. For each text instance, we use
\begin{equation}\label{eq:regress}
   \Delta_{x} = b_{ix}-x_{min},
   \;\;\;
   \Delta_{y} = b_{iy}-y_{min},
 \end{equation}
where $x_{min}$ and $y_{min}$ represent the minimum $x$ and $y$ values of the 4 vertexes, respectively. The advantage of predicting the relative distance is that it is irrelevant to whether the Bezier curve control points are beyond the image boundary. Inside the detection head, we only use one convolution layer with $4(n+1)$ ($n$ is the number of Bezier Curve order) output channels to learn the $\Delta_{x}$ and $\Delta_{y}$, which is nearly cost-free while the results can still be accurate.
We discuss the details in \S\ref{sec:exp}.

\subsubsection{Bezier Ground-truth Generation}
\label{subsubsec:Bezier_gen}
\begin{figure*}[!t]
   \centering
   \centerline{\includegraphics[width=17.6cm, height = 4cm]{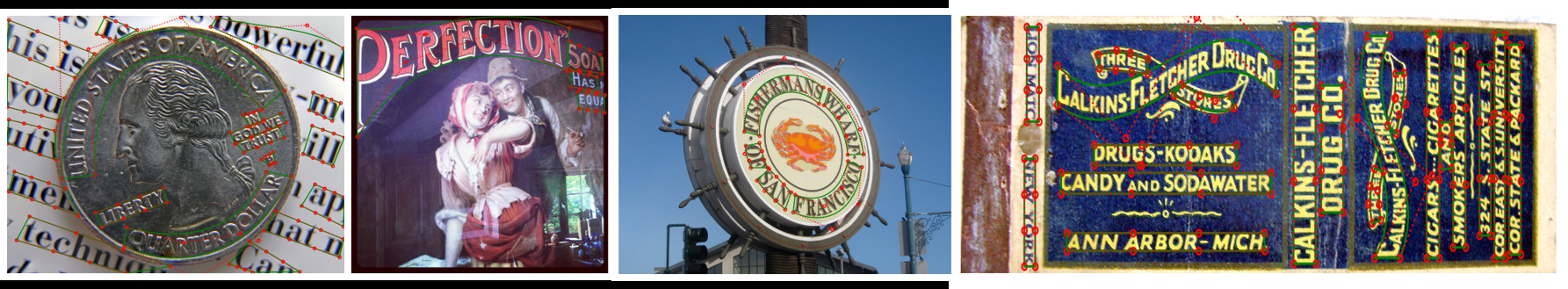}}
   \caption{
   \textbf{Example results of Bezier curve generation}. Green lines are the final Bezier curve results. Red dash lines represent the control polygon, and the 4 red end points represent the control points. Zoom in for better visualization.}\label{fig:Bezier_gen_examples}
\end{figure*}

In this section, we briefly introduce how to generate Bezier curve ground-truth based on the original annotations. The arbitrarily-shaped datasets, \textit{e.g.}, Total-text \cite{ch2019total} and SCUT-CTW1500 \cite{liu2019curved}, use polygonal annotations for the text regions. Given the annotated points $\{p_{i}\} _{i=1}^{n}$ from the curved boundary, where $p_{i}$ represents the $i$-$th$ annotating point, the main goal is to obtain the optimal parameters for cubic Bezier curves $c(t)$ in Equation \eqref{eq:Bezier_def}. To achieve this, we can simply apply the standard least-squares fitting as follows:
\begin{equation}
\begin{bmatrix}
  B_{0,n}(t_0) & \cdots\ & B_{n,n}(t_0)\\
  B_{0,n}(t_1) & \cdots\ & B_{n,n}(t_1)\\
  \vdots &  \ddots & \vdots\\
  B_{0,n}(t_m) & \cdots\ & B_{n,n}(t_m)
\end{bmatrix}
\begin{bmatrix}
 b_{x_0} & b_{y_0}\\
 b_{x_1} & b_{y_1}\\
 \vdots  & \vdots \\
 b_{x_n} & b_{y_n}\\
\end{bmatrix}
=
\begin{bmatrix}
 p_{x_0} & p_{y_0}\\
 p_{x_1} & p_{y_1}\\
 \vdots & \vdots\\
 p_{x_m} & p_{y_m}\\
\end{bmatrix}
.
\label{eq:matrix}
\end{equation}
Here $m$ represents the number of the annotated points for a curved boundary. For Total-Text and SCUT-CTW1500, $m$ is 5 and 7, respectively. $t$ is calculated by using the ratio of the cumulative length to the perimeter of the poly-line. According to Equation \eqref{eq:Bezier_def} and Equation \eqref{eq:matrix}, we convert the original poly-line annotation to a parameterized Bezier curve. Note that we directly use the first and the last annotating points as the first ($b_0$) and the last ($b_n$) control points, respectively. An visualization comparison is shown in Figure \ref{fig:Bezier_gen_comp}, which shows that the generated results can be even visually better than the original annotation.
In addition, thanks to the structured output, the task of text recognition can be easily formulated by applying our proposed BezierAlign (see \S\ref{subsec:bezieralign}), which warps the curved text into horizontal representation. More results of the Bezier curve generation are shown in Figure \ref{fig:Bezier_gen_examples}. The simplicity of our method allows it to deal with various shapes in a unified representation format.

\begin{figure*}[t!]
   \begin{minipage}[c]{0.33\linewidth}
     \centering
     \centerline{\includegraphics[width = 5.5cm, height = 3.5cm]{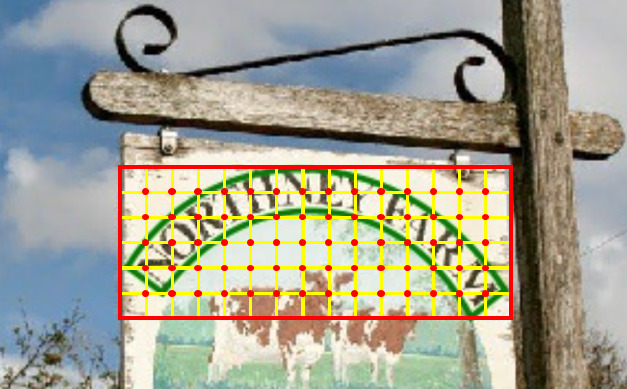}}\label{fig:det_c1}
     \centerline{\small{(a) Horizontal sampling. }}\medskip
   \end{minipage}
   \begin{minipage}[c]{0.33\linewidth}
     \centering
     \centerline{\includegraphics[width = 5.5cm, height = 3.5cm]{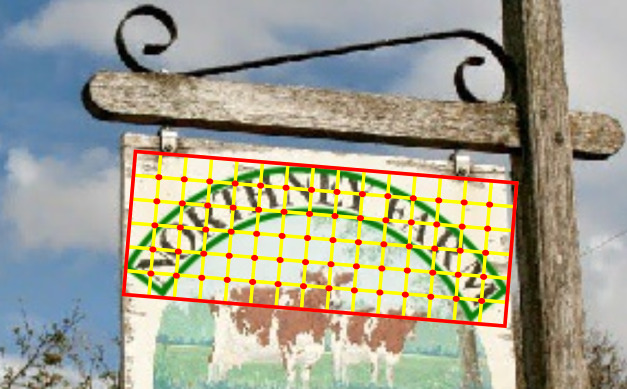}}\label{fig:det_c2}
     \centerline{\small{(b) Quadrilateral sampling. }}\medskip
   \end{minipage}
   \begin{minipage}[c]{0.33\linewidth}
     \centering
     \centerline{\includegraphics[width = 5.5cm, height = 3.5cm]{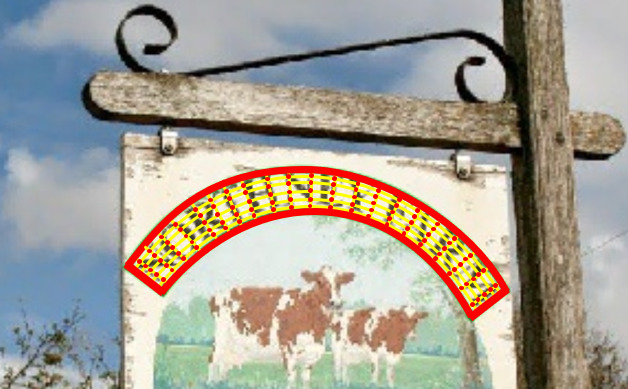}}\label{fig:det_c3}
     \centerline{\small{(c) BezierAlign. }}\medskip
   \end{minipage}

   \caption{
   \textbf{Comparison between previous sampling methods and BezierAlign}. The proposed BezierAlign can more accurately sample features of the text region, which is essential for achieving good recognition accuracy.
   Note that the alignment procedure is applied to intermediate convolution features.}\label{fig:Bezier_align}
\end{figure*}

\subsection{CoordConv}
\label{subsec:coordconv}
As pointed out in \cite{liu2018intriguing}, conventional convolutions show limitation when learning a mapping between coordinates in $(x, y)$
Cartesian space and coordinates in one-hot pixel space. The problem can be effectively solved by concatenating the coordinates to the feature maps. The recent practice of encoding relative coordinates \cite{wang2020solov2} also show that the relative coordinates can provide informative cues for instance segmentation.

Let $f_{outs}$ denotes the features of different scales of FPN, and $O_{i,x}$ and $O_{i,y}$ represent the absolute $x$ and $y$ coordinates, respectively, from all the locations (\textit{i.e.}, the location where the filters are generated) for the $i^{th}$ level of FPN. All $O_{i,x}$ and $O_{i,y}$ consist of two feature maps $f_{ox}$ and $f_{oy}$. We simply concatenate $f_{ox}$ and $f_{oy}$ to the last channel of $f_{outs}$ along the channel dimension. Therefore, new features $f_{coord}$ with additional two channels are formed, which are subsequently input to three convolutional layers with kernel size, stride, and padding size setting to 3, 1, and 1, respectively.
We find that using such simple coordinate convolutions can considerably
improve the performance of scene text spotting.

\subsection{BezierAlign}
\label{subsec:bezieralign}
To enable end-to-end training, most of the previous methods adopt various sampling (feature alignment) methods to connect the recognition branch.
Typically, a sampling method represents an in-network region cropping procedure. In other words, given a feature map and Region-of-Interest (RoI), using the sampling method to extract the features of RoI and efficiently output a feature map of a fixed size.
However, sampling methods of previous non-segmentation based methods, \textit{e.g.}, RoI Pooling \cite{li2017towards}, RoI-Rotate \cite{liu2018fots}, Text-Align-Sampling \cite{he2018end}, or RoI Transform \cite{sun2018textnet} cannot properly align features of arbitrarily-shaped text. By exploiting the parameterization nature of a structured Bezier curve bounding box, we propose BezierAlign for feature sampling/alignment, which may be viewed as a flexible extension of
RoIAlign \cite{He2017Mask}.
Unlike RoIAlign, the shape of the sampling grid of BezierAlign is not rectangular. Instead, each column of the arbitrarily-shaped grid is orthogonal to the Bezier curve boundary of the text. The sampling points have equidistant interval in width and height, respectively, which are bilinear interpolated with respect to the coordinates.

Formally, given an input feature map and Bezier curve control points, we
process all the output pixels of the rectangular output feature map with size $h_{out}\times w_{out}$. Taking pixel $g_i$ from output feature map with position $(g_{iw}, g_{ih})$ as an example, we calculate $t$
as follows:
\begin{equation}\label{eq:Bezier_timestep}
   t = \frac{g_{iw}}{w_{out}}.
\end{equation}
Then the points of upper Bezier curve boundary $tp$ and lower Bezier curve boundary $bp$ are calculated according to Equation \eqref{eq:Bezier_def}. Using $tp$ and $bp$, we can linearly index the sampling point $op$ by Equation \eqref{eq:Bezier_sampling}:
\begin{equation}\label{eq:Bezier_sampling}
   op = bp  \cdot  \frac{g_{ih}}{h_{out}} + tp \cdot  (1 - \frac{g_{ih}}{h_{out}}).
\end{equation}
With the position of $op$, we can easily apply bilinear interpolation to calculate the result. Due to the accurate sampling of features, the performance of text recognition is improved substantially. We compare BezierAlign with other sampling strategies, as presented in Figure \ref{fig:Bezier_align}.

\subsection{Attention-based Recognition Branch}
Benefiting from the shared backbone features and BezierAlign, we design a light-weight recognition branch as shown in Table \ref{tab:rec_struct}, for faster execution. It consists of 6 convolutional layers, 1 bidirectional LSTM layer, and an attention-based recognition module. In the conference version \cite{liu2020abcnet}, we have applied %
the
CTC loss \cite{graves2006connectionist} for text string alignment between predictions and ground-truth, but we find that the attention-based recognition module \cite{bahdanau2014neural,donahue2015long,liao2019mask,wang2020decoupled} is more powerful and can lead to better results.
In the inference phase, the RoI region is replaced by the detected  Bezier curve as in \S\ref{subsec:Bezier_det}. Note that in
\cite{liu2020abcnet}, we only use the generated Bezier curves to extract the RoI features during training. In this paper, we also take advantages of the detection results (see \S\ref{sec:aet}).

The attention mechanism takes zero RNN initial states and the embedding features of an initial symbol for the sequential prediction. During each step, the $c$-category softmax prediction (representing the predicted character), previous hidden state, and a weighted sum of the cropped Bezier curved features are recursively used to compute the results. The prediction continues until an End-of-Sequence (EOS) symbol is predicted.
The number of the class is set to 96 (excluding the EOS symbol) for English, while for the bilingual task including Chinese and English, the number of the class is set to 5462. Formally, at time step $t$, the attention weights are calculated by:
\begin{equation}
  e_{t,s} = \pmb{K}^{T} { \rm tanh}(\pmb{W}h_{t-1}+\pmb{U}h_{s}+\pmb{b}),
\end{equation}
where, $h_{t-1}$ is the last hidden state, $\pmb{K}$, $\pmb{W}$, $\pmb{U}$, and $\pmb{b}$ are the learnable weight matrices and parameters. The weighted sum of the sequential feature vectors is formulated as:
\begin{equation}
  c_t = \sum_{s=1}^{n}a_{t,s}h_{s},
\end{equation}
where $a_{t,s}$ is defined as:
\begin{equation}
  a_{t,s} = \frac{ \exp(e_{t,s})}{\sum_{s=1}^{n}  \exp(e_{t,s})}.
\end{equation}
Then, the hidden state can be updated, as follows:
\begin{equation}
  h_t = { \rm GRU} ((\pmb{embed}_{t-1}, c_t), h_{t-1}).
\end{equation}
Here
$\pmb{embed}_{t-1}$ is an embedding vector of the previous decoding result $y_t$, which is generated by the classifier:
\begin{equation}
  y_t = \pmb{w}h_t + \pmb{b}.
\end{equation}
Therefore, we %
use the softmax function to estimate the probability distribution $p(u_t)$.
\begin{equation}
  u_t = {\rm softmax}(\pmb{V}^{T}h_{t}),
\end{equation}
where $\pmb{V}$ represents the parameters to be learned. To stabilize training, we also use a teacher enforcing strategy \cite{williams1989learning}, which delivery a ground-truth character instead of the prediction of the GRU for the next prediction under a predefined probability setting to 0.5 in our implementation.

\begin{table}[!t]
   \centering
   \newcommand{\tabincell}[2]{\begin{tabular}{@{}#1@{}}#2\end{tabular}}
   \caption{Structure of the recognition branch, which is a simplified version of CRNN \cite{shi2017end}. For all convolutional layers, the padding size is restricted to 1. $n$ represents batch size. $c$ represents the channel size. $h$ and $w$ represent the height and width of the outputted feature map, and $n_{class}$ represents the number of the predicted class.}
   \label{tab:rec_struct}
   \footnotesize
   \begin{tabular}{c|c|c}
     \hline
     \tabincell{c}{Layers \\ (CNN - RNN)}  & \tabincell{c}{Parameters \\ (kernel size, stride)}  & \tabincell{c}{Output  Size \\ ($n$, $c$, $h$, $w$)} \\
     \hline
     conv.\ layers $\times$4 & (3, 1) & ($n$, 256, $h$, $w$) \\
     conv.\  layers $\times$2 & (3, (2,1)) & ($n$, 256, $h$, $w$) \\
     average pool for $h$ &  -  & ($n$, 256, 1, $w$) \\
     \hline
     Channels-Permute & - & ($w$, $n$, 256) \\
     BLSTM  & - & ($w$, $n$, 512) \\
     Attention-based decoder  & - & ($w$, $n$, $n_{class}$) \\
     \hline
   \end{tabular}
 \end{table}

\subsection{Adaptive End-to-End Training}
\label{sec:aet}
In our published conference version \cite{liu2020abcnet}, we only use
the ground truth to BezierAlign for the text recognition branch during the training phase. While in the testing phase, we use the detection results for feature cropping. Based on the observations, some errors may occur when the detection results are not as accurate as the ground-truth Bezier curved bounding box. To alleviate such issues, we propose a simple yet effective strategy, termed Adaptive End-to-End Training (AET).

Formally, the detection results are first suppressed by confidence thresholds, and then using NMS to eliminate the redundant detection results. The corresponding recognition ground truth is then assigned to each detection result based on the minimum summation of the distances of the coordinates of control points:
\begin{equation}
  rec = \mathop{\arg\min}_{rec^*\in cp^*}\sum_{i=1}^{n} |cp_{x_i,y_i}^* - cp_{x_i,y_i}|,
\end{equation}
where $cp_*$ is the ground truth of the control points. $n$ is the number of the control points. After assigning the recognition annotation to the detection results, we simply concatenate the new target to the original ground truth set for further recognition training.

\subsection{Text Spotting Quantization}
Scene text reading applications usually require a real-time performance; however, there are few works attempting to use quantization for scene text spotting task.
Model quantization aims at discretizing the full precision tensors into low-bit tensors without much degradation of network performance.
Limited number of representation levels (quantization levels) are available. Given quantization bit width to be $b$-bit, the number of quantization levels are $2^b$. It is easy to see that deep learning models might suffer significant performance drop with the quantization bit width becoming low.
To maintain the accuracy, the discretization error should be minimized:
\begin{equation}
\label{eq:alg-1}
Q^\ast(x) = \mathop{\arg\min}_{Q} \sum (Q(x) -x)^2,
\end{equation}
where $Q(x)$ is the quantization function. Motivated by the LSQ \cite{esser2019learned}, we employ the following equations as the activation quantifiers in this paper. Specifically, for any data $x^a$ from the activation tensor $X^a$, its quantization value $Q(x^a)$ is computed by a serial of transformations.

\begin{figure*}[t!]
\centering
\begin{minipage}[c]{.3\linewidth}
  \includegraphics[width = 5.5cm, height = 3.5cm]{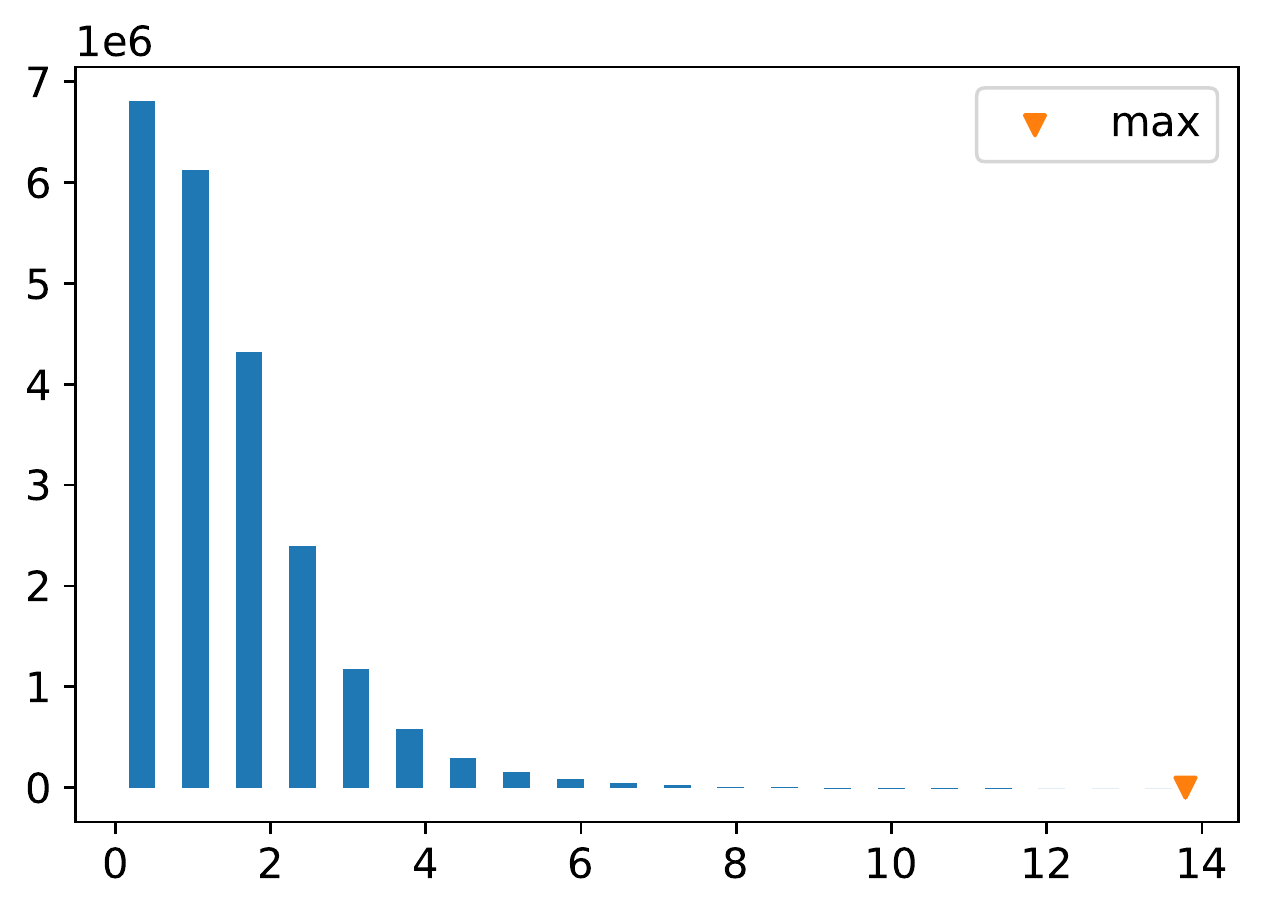}
  \centerline{\small{(a) 1st layer activation}}
\end{minipage}
\begin{minipage}[c]{.3\linewidth}
  \includegraphics[width = 5.5cm, height = 3.5cm]{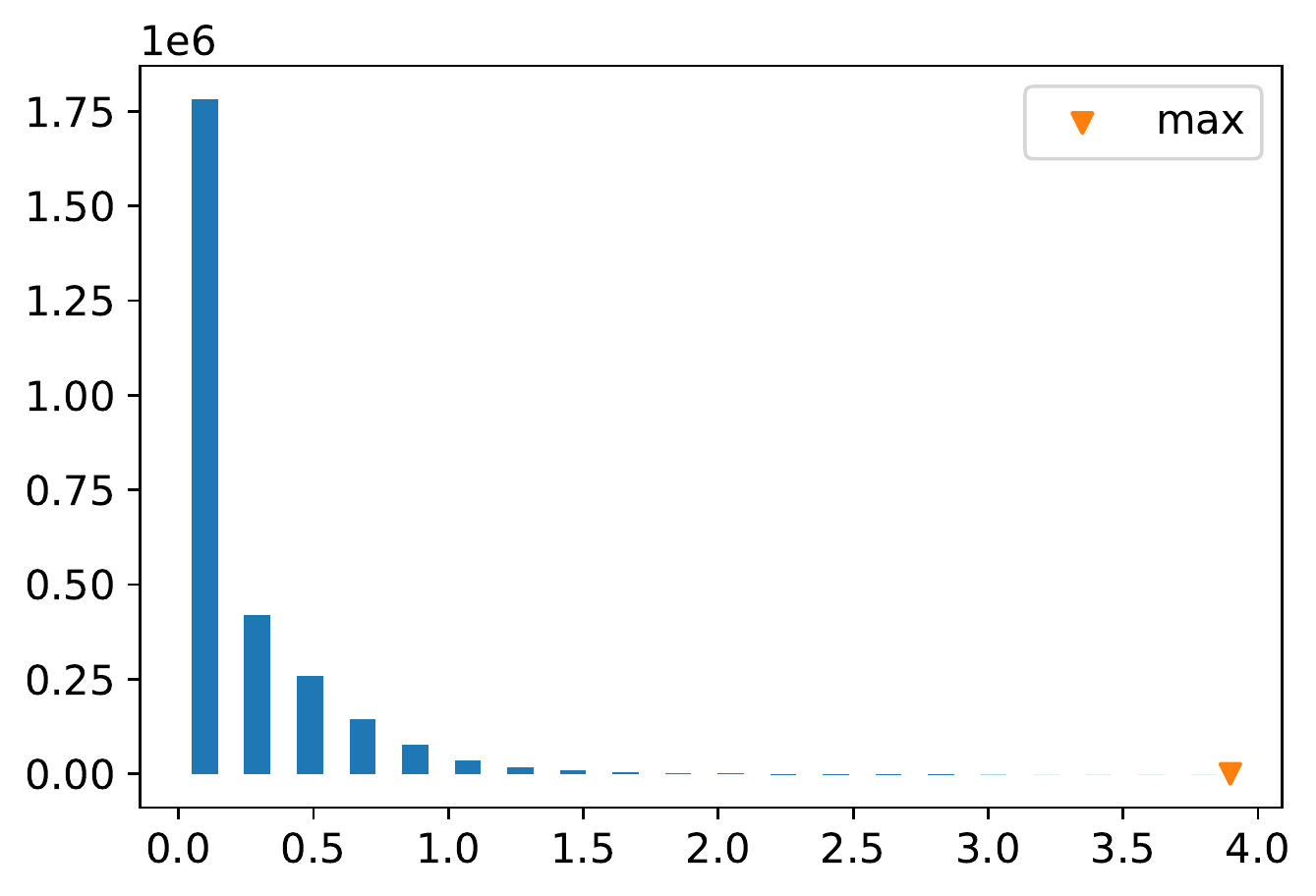}
  \centerline{\small{(b) 21st layer activation}}
\end{minipage}
\begin{minipage}[c]{.3\linewidth}
  \includegraphics[width = 5.5cm, height = 3.5cm]{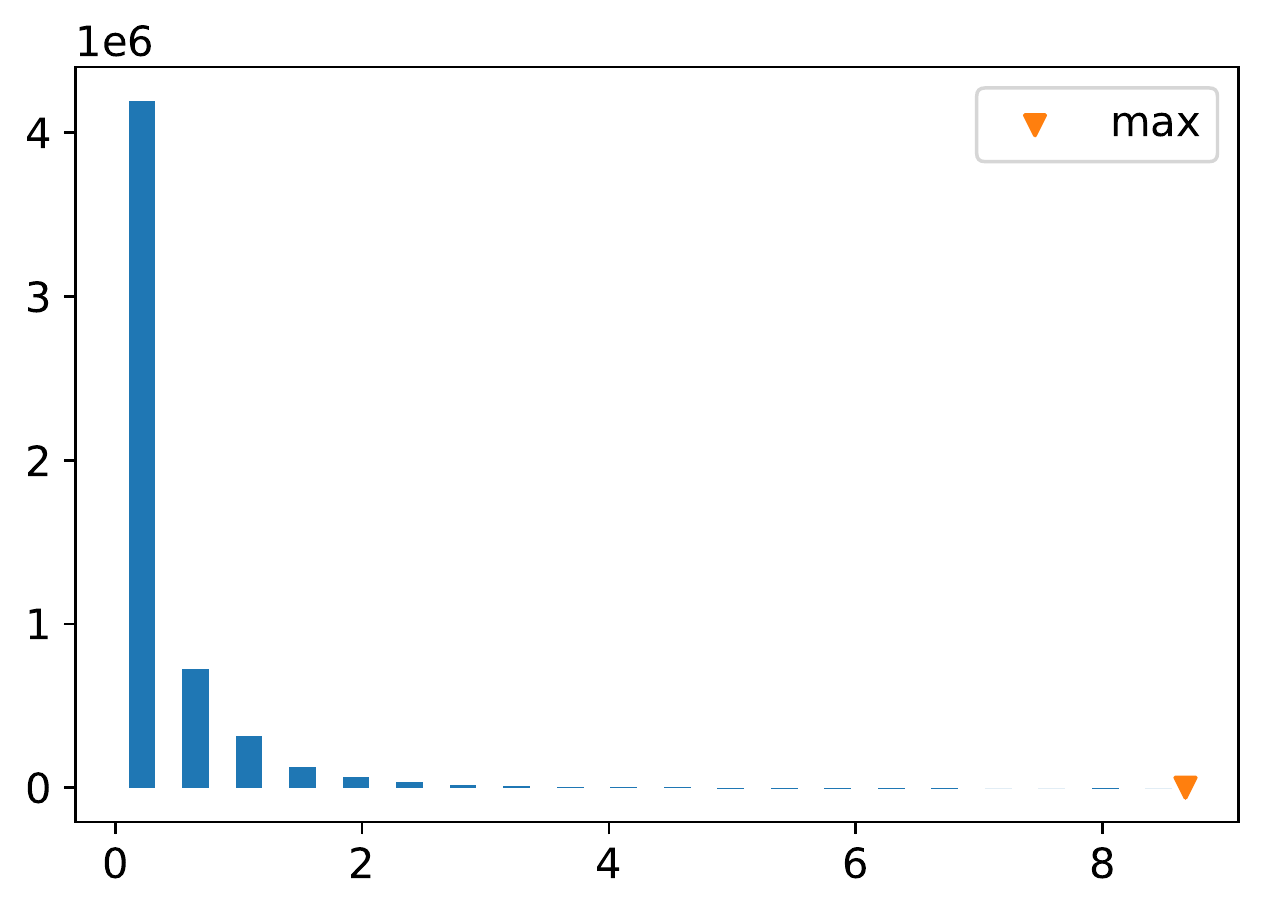}
  \centerline{\small{(c) 31st layer activation}}
\end{minipage}

\caption{\textbf{Data distribution in the ABCNet v2 with the max value marked in the figure.}
We can see that the abnormal large values have a low occurrence frequency.}
\label{fig:abnormal}
\end{figure*}

Firstly, as indicated in the work PACT \cite{choi2018pact}, not all the full precision data should be linearly mapped to the quantized value. It is common to find some abnormal large values, which rarely occur in the full precision tensor. We further visualize data distribution of some layers from the ABCNet v2 in Figure \ref{fig:abnormal} and observe a similar phenomenon. Therefore, a learnable parameter $\alpha^a$ is assigned to dynamically depict the discretization range, with beyond data clipped to the boundary:
\begin{equation}
\label{eq:quant-1}
y^a = {\rm min}({\rm max}(x^a, 0), \alpha^a),
\end{equation}
Secondly, data in the clipped range (so-called quantization interval) is linearly mapped to nearby integer, as shown in Equation~\eqref{eq:quant-2}:
\begin{equation}
\label{eq:quant-2}
z^a = \lfloor \frac{y^a}{\alpha^a} \cdot (l-1) \rceil ,
\end{equation}
where $l=2^b$ is the number of quantization levels mentioned above and $\lfloor \cdot \rceil$ is the nearest-rounding function.
Thirdly, to keep the data magnitude similar before and after quantization, we apply corresponding scale factor on $z^a$ to obtain $Q(x^a)$ by:
\begin{equation}
\label{eq:quant-3}
Q(x^a) = z^a \cdot \frac{\alpha^a}{l-1},
\end{equation}
In summary, the quantization of activations can be written as:
\begin{equation}
\label{eq:quant-4}
\begin{aligned}
Q(x^a) &= \lfloor {\rm min}({\rm max}(x^a, 0), \alpha^a) \cdot \frac{l-1}{\alpha^a} \rceil \cdot \frac{\alpha^a}{l-1} \\
 &= \lfloor {\rm min}({\rm max}(\frac{x^a}{\alpha^a}, 0), 1) \cdot (l-1) \rceil \cdot \frac{\alpha^a}{l-1},
\end{aligned}
\end{equation}
Different from activations, weight parameters generally contain both positive and negative values, thus extra linear transforms are introduced before discretization as follows:
\begin{equation}
\label{eq:quant-5}
\begin{aligned}
z^w &=  \lfloor ({\rm min}({\rm max}(\frac{x^w}{\alpha^w}, -1), 1) + 1) / 2 \cdot (l-1) \rceil,\\
Q(x^w) &= (\frac{z^w}{l-1} \cdot 2 - 1) \cdot \alpha^w
=  (2\cdot z^w - (l-1)) \cdot \frac{\alpha^w}{l-1}\\
\end{aligned}
\end{equation}
%
%
%
One issue for model quantization is the gradient vanishing caused by the $\rm round$ function ($\lfloor \cdot \rceil$). This is because the $\rm round$ function has an almost-everywhere zero gradient. Straight through estimator (STE) is employed to tackle the problem. Specifically, we override the derivative of the $\rm round$ function constantly to be 1 ($\partial{\lfloor \cdot \rceil} = 1$). We employ mini-batch gradient descent optimizer to train the quantization related parameters $\alpha^a$ and $\alpha^w$ in each layer together with the original parameters from the network.

It should be noted that for each convolutional layer, its quantization introduced parameter $\alpha^a$ and $\alpha^w$ are shared for all elements in the input activation tensor $X^a$ and weight tensor $X^w$, respectively. Thus, it is possible to exchange the computational order during network forward-propagation, as depicted in Equation~\ref{eq:quant-7}, for better efficiency.
With the exchange, the time-consuming convolutional computation is operated in integer format only (all elements $z^a \in Z^a$ and $z^w \in Z^w$ are $b$-bit integers).
Therefore, benefits in aspects of latency, memory footprint and energy consumption can be achieved compared with corresponding floating-point counterpart.
\begin{equation}
\label{eq:quant-7}
Q(X^a) \cdot Q(X^w) = (Z^a \cdot (2 \cdot Z^w - (l-1))) \cdot (\frac{\alpha^a \cdot \alpha^w}{(l-1)^2}),
\end{equation}

In theory, for $b$-bit quantization network, the input activation and weight have a $\frac{32}{b} \times$ memory saving. For energy consumption, we list the estimated energy cost for per operation of different types on chip in Table \ref{tab:hardware}. As we can see, the energy cost of floating-point $\rm ADD$ and $\rm MULT$ is much larger than that of the fixed-point operation. Moreover, the DRAM access costs magnitude-order higher energy compared to the ALU operations. Therefore, it is clear that the quantized model is potential to save considerable energy compared to the full precision counterpart.
\begin{table}[b!]
\centering
\caption{Energy consumption of different operations in 45nm CMOS process.}
\label{tab:hardware}
\begin{tabular}{ l | l }
\hline
Operation & Energy (pJ) \\
\hline
32-bit Fixed-point ADD & 0.1 \\
32-bit floating-point ADD & 0.9 \\
32-bit Fixed-point MULT & 3.1 \\
32-bit floating-point MULT & 3.7 \\
32-bit 32KB SRAM & 5 \\
32-bit DRAM & 640 \\
\hline
\end{tabular}
\end{table}

In terms of inference latency, the actual speedup of quantized models against full precision counterparts is decided by the computational ability of fixed-point arithmetic versus floating-point arithmetic on the platform. Table \ref{tab:nvidia} shows the operations per cycle per SM on the Nvidia Turing architecture. We can learn from Table \ref{tab:nvidia} that an 8-bit network is potential to achieve 2$\times$ speedups versus the full precision counterpart on the platform. More impressively, 4-bit network and binary neural network (1-bit) are able to run faster than the full precision model by 4$\times$ and 16$\times$, respectively.

\begin{table}[hbt!]
\centering
\caption{Computational ability (Ops per cycle per SM) comparison on Nvidia Turing Architecture.}
\label{tab:nvidia}
\begin{tabular}{ l | l | l }
\hline
input precision & Output & Ops/Cycle/SM\\
\hline
FP16 & FP16 or FP32 & 1024 \\
INT8 & INT32 &  2048 \\
INT4 & INT32 & 4096 \\
INT1 & INT32 & 16384\\
\hline
\end{tabular}
\end{table}

\section{Experiments}
\label{sec:exp}
To evaluate the effectiveness of ABCNet v2, we conduct experiments on various scene text benchmarks, including multi-oriented scene text benchmarks ICDAR'15 \cite{karatzas2015icdar}, MSRA-TD500 \cite{yao2012detecting}, ReCTS \cite{zhang2019icdar}, and two arbitrarily shaped benchmarks Total-Text \cite{ch2019total} and SCUT-CTW1500 \cite{liu2019curved}. The ablation studies are conducted on Total-Text and SCUT-CTW1500 to verify each component of our proposed method.

\subsection{Implementation  Details}
\label{subsec:Imple_details}
The backbone of the work here follows a common setting as most previous work, \textit{i.e.}, ResNet-50 \cite{he2016identity} together with a Feature Pyramid Network (FPN) \cite{lin2017feature} unless specified otherwise. For the detection branch, we apply RoIAlign on 5 feature maps with 1/8, 1/16, 1/32, 1/64, and 1/128 resolution of the input image while for recognition branch, BezierAlign is conducted on three feature maps with 1/4, 1/8, and 1/16 sizes, and the width and height of the sampling grid are set to 8 and 32, respectively.
For English only dataset, the pretrained data is collected from publicly available English word-level-based datasets, including 150K synthesized data described in the next section, 7K ICDAR-MLT data \cite{nayef2019icdar2019}, and the corresponding training data of each dataset. The pretrained model is then fine-tuned on the training set of the target datasets. Note that 15k COCO-Text \cite{veit2016coco} images in our previous manuscript \cite{liu2020abcnet} are not used in this improved version. For ReCTS dataset, we adopt LSVT \cite{sun2019icdar}, ArT \cite{chng2019icdar2019}, ReCTS \cite{zhang2019icdar}, and the synthetic pretrained data to train the model.

In addition, we also adopt data augmentation strategies, \textit{e.g.}, random scale training, with the short size uniquely chosen from 640 to 896 (interval of 32) and the long size being less than 1600; and random crop, which we make sure that the crop image do not cut the text instance (for some special cases that hard to meet the condition, we do not apply random crop).

We train our model using 4 Tesla V100 GPUs with the image batch size of 8. The maximum iteration is 260K; and the initialized learning rate is 0.01, which reduces to 0.001 at the 160K$^{\rm th}$ iteration and 0.0001 at 220K$^{\rm th}$ iteration.

\begin{figure*}[!ht]
  \centering
  \centerline{\includegraphics[
  width=0.96\textwidth
  ]{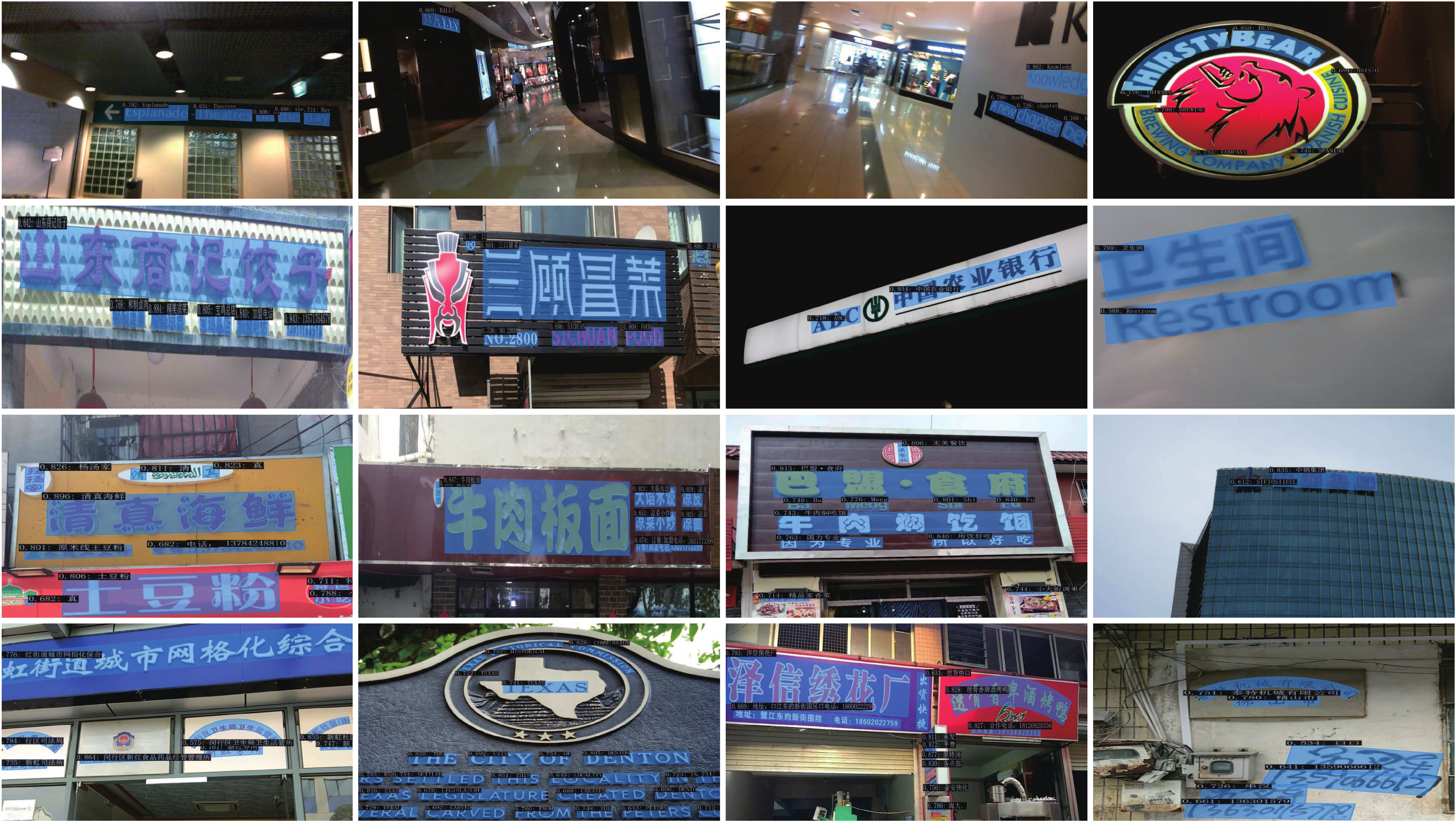}}
  \caption{\textbf{Qualitative results of ABCNet v2 on various datasets.} The detection results are shown with blue bounding boxes.
  Prediction confidence scores
  are also shown. Best viewed on screen.
  }\label{fig:FIRES}
\end{figure*}

\subsection{Benchmarks}
{\textbf Bezier Curve Synthetic Dataset 150k.} For the end-to-end scene text spotting methods, a massive amount of free synthesized data are
always necessary. However, the existing 800k SynText dataset \cite{gupta2016synthetic} only provides a quadrilateral bounding box for a majority of straight text. To diversify and enrich the arbitrarily shaped scene text, we make some effort to synthesize a dataset of 150K images (94,723 images contain a majority of straight text, and 54,327 images contain mostly curved text) with the VGG synthetic method \cite{gupta2016synthetic}.

\begin{figure}[!b]
   \centering
   \centerline{\includegraphics[
   width=0.48\textwidth
   ]{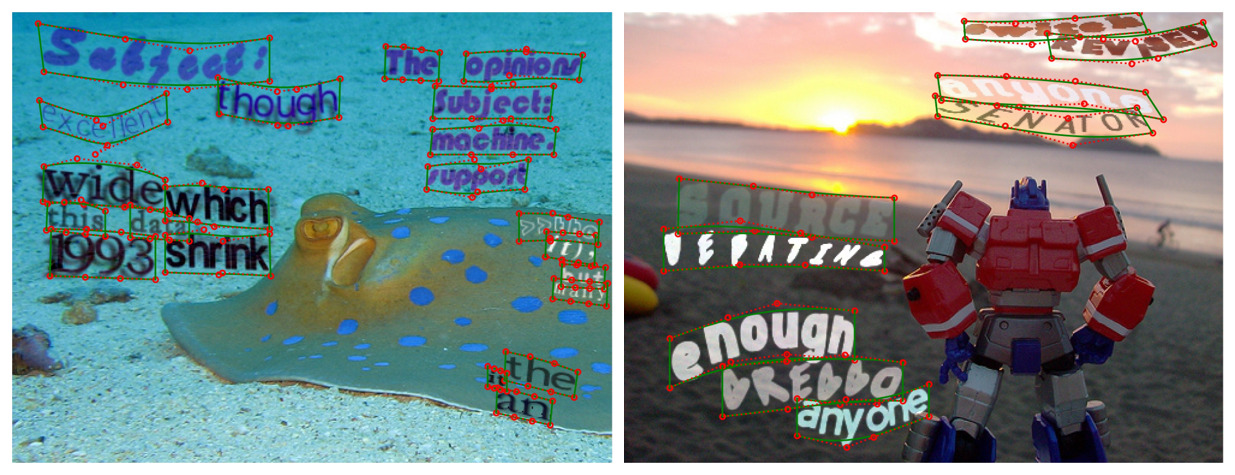}}
   \caption{\textbf{Examples of English Bezier curve synthesized data.}}\label{fig:Bezier_syn}
\end{figure}

\begin{figure}[!h]
  \centering
  \centerline{\includegraphics[
  width=0.48\textwidth
  ]{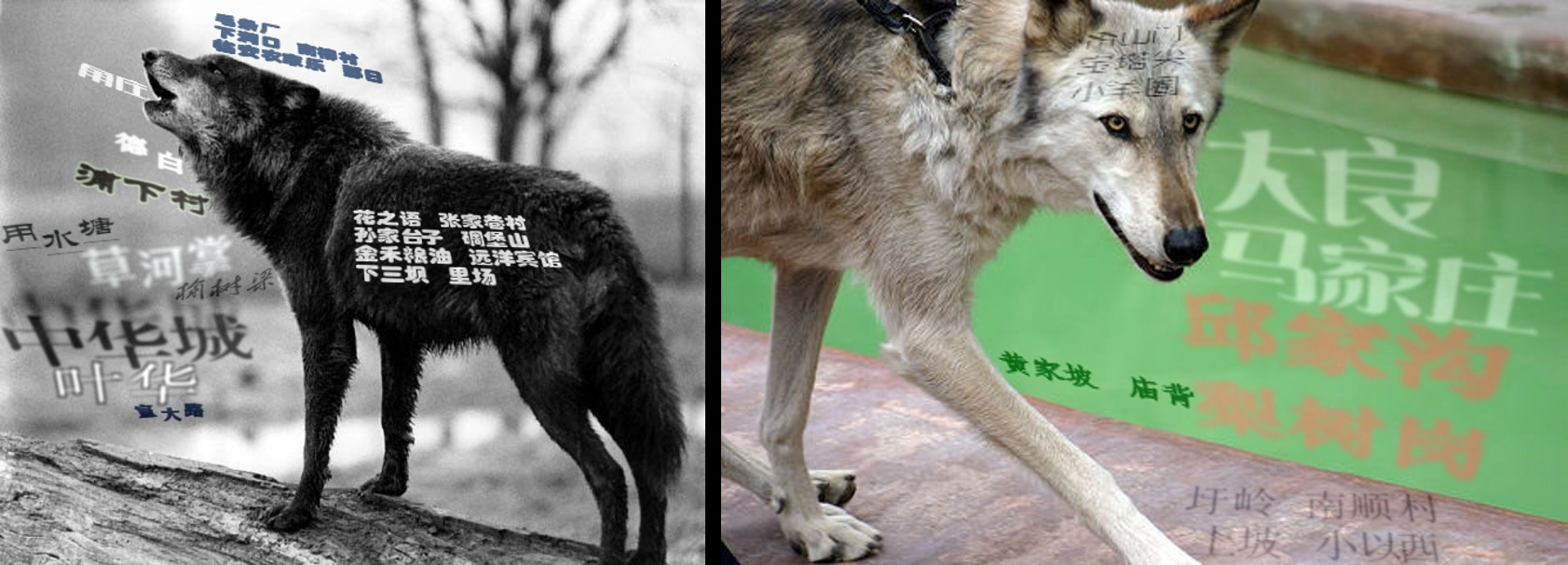}}
  \caption{\textbf{Examples of Chinese Bezier curve synthesized data.}}\label{fig:Bezier_syn_chn}
\end{figure}

Specially, we filter out 40K text-free background images from COCO-Text \cite{veit2016coco} and then prepare the segmentation mask and scene depth of each background image for the following text rendering. To enlarge the shape diversity of synthetic texts, we modify the VGG synthetic method by synthesizing scene text with various art fonts and corpus and generate the polygonal annotation for all the text instances. The annotations are then used for producing the Bezier curve ground truth by the generating method described in \S\ref{subsubsec:Bezier_gen}. Examples of our synthesized data are shown in Figure \ref{fig:Bezier_syn}. For Chinese pretraining, we synthesized 100K images following the same method as above, with some examples shown in Figure \ref{fig:Bezier_syn_chn}.

{\bf Total-Text} dataset \cite{ch2019total} is one of the most important arbitrarily shaped scene text benchmark proposed in 2017, which was collected from various scenes, including text-like scene complexity and low-contrast background. It contains 1,555 images, with 1,255 for training and 300 for testing. To resemble the real-world scenarios, most of the images of this dataset contain a large amount of regular text, while guarantee that each image has at least one curved text. The text instance is annotated with polygon based on word-level. Its extended version \cite{ch2019total} improves its annotation of the training set by annotating each text instance with a fixed ten points following text recognition sequence. The dataset contains English text only. To evaluate the end-to-end results, we follow the same metric as previous methods, which use F-measure to measure the word-accuracy.

{\bf SCUT-CTW1500} dataset \cite{liu2019curved} is another important arbitrarily shaped scene text benchmark proposed in 2017. Compared to Total-Text, this dataset contains both English and Chinese text. In addition, the annotation is based on text-line level, and it also includes some document-like text, i.e., numerous small text may stack together. SCUT-CTW1500 contains 1k training images, and 500 testing images.

{\bf ICDAR 2015} dataset \cite{karatzas2015icdar} provides images which are incidentally captured in the real world. Unlike previous ICDAR datasets, in which the text are clean, well-captured, and horizontally centered in the images. The dataset includes 1000 training images and 500 testing images with complicate backgrounds. Some text may also appear in any orientation and any location, with small size or low resolution. The annotation is based on word-level, and it only includes English samples.

{\bf MSRA-TD500} dataset \cite{yao2012detecting} contains 500 multi-oriented Chinese and English images, with 300 images for training and 200 images for testing. Most of the images are captured indoor. To overcome the insufficiency of the training data, we use the synthetic Chinese data mentioned above for model pretraining.

{\bf ICDAR'19-ReCTs} dataset \cite{zhang2019icdar} contains 25k annotated signboard images, in which 20k images are for training set, and the rest are for testing set. Compared to English text, Chinese text normally has a significantly large number of the classes, with more than 6k commonly used characters with complicated layouts and various fonts. This dataset mainly contains text of the shop signs, and it also provides annotations for each character.

{\bf ICDAR'19-ArT} dataset \cite{chng2019icdar2019} is currently the largest dataset for arbitrarily shaped scene text. It is the combination and extension of the Total-text and SCUT-CTW1500. The new images also contains at least one arbitrarily-shaped text per image. There is a high diversity in terms of text orientations. The ArT dataset is split to a training set with 5,603 images and 4,563 for testing set. All the English and Chinese text instances are annotated with tight polygons.

{\bf ICDAR'19-LSVT} dataset \cite{sun2019icdar} provides an unprecedentedly large number of text from street view. It provides total 450k images with rich information of the real scene, among which 50k are annotated in full annotations (30k for training and the rest 20k for testing). Similar to ArT \cite{chng2019icdar2019}, this dataset also contains some curved text, which are annotated with polygon.

\subsection{Ablation Study}
To evaluate the effectiveness of the proposed components, we conduct ablation studies on two datasets, Total-Text and SCUT-CTW1500. We
find that there are some training errors because of the different initialization. In addition, the text spotting task requires that all the characters should be correctly recognized. To avoid such issues, we train each method for three times and report the average results. The results are shown in Table \ref{tab:ablative}, which demonstrate that all modules can lead to an obvious improvement against the baseline model on both datasets.

We can see that using the attention-based recognition module, the results can be improved by 2.7\% on Total-Text and 7.9\% on SCUT-CTW1500, respectively.

\begin{figure}[!b]
  \centering
  \centerline{\includegraphics[
  width=0.48\textwidth
  ]{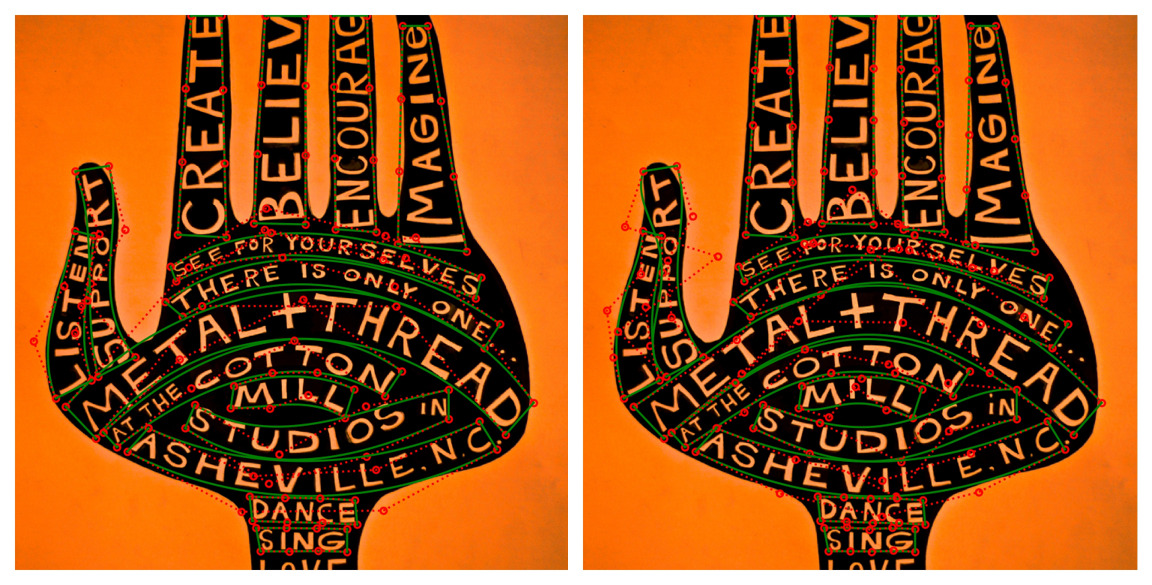}}
  \caption{\textbf{Comparison between cubic Bezier curve (left) and 4th-order Bezier curve (right).} There are some slight differences.}\label{fig:4order}
\end{figure}

We then evaluate all other modules using the attention-based recognition branch. Some conclusions are as follows:
\begin{itemize}
    \item

Using a biFPN architecture, the results can be improved by an additional 1.9\% and 1.6\%, while the inference speed is only reduced by 1 FPS.
Thus, we achieve a better trade-off between speed and accuracy.

\item

Using coordinate convolution mentioned in \S\ref{subsec:coordconv}, the results can be significantly improved by 2.8\% and 2.9\% on two datasets, respectively. Note that such improvement does not introduce noticeable
computation overhead.

\item
We also test the AET strategy mentioned in \S\ref{sec:aet}, which results in 1.2\% and 1.7\% improvement.

\item
Lastly, we conduct experiments to show how the setting of Bezier curve order affects the results. Specifically, we regenerate all the ground truths for the same synthetic and the real images using 4th-order Bezier curves. We then train the ABCNet v2 by regressing the control points and use 4th-order BezierAlign. Other parts remain the same as the 3rd-order setting. The results shown in Table \ref{tab:ablative} demonstrate that increasing the order can be conducive to the text spotting results, especially on SCUT-CTW1500 which adopts text-line annotation.
We further conduct experiments by using 5th-order Bezier curves on Total-text dataset following the same experimental setting; however, compared with baseline, we find that the performance drops from 66.2\% to 65.5\% in terms of the E2E Hmean. Based on the observation, we assume that the decline might be because an extremely higher order may result in drastic variation of the controlled points, which could exacerbate the difficulties of the regression.
Some results using 4th-order Bezier curve are shown in Figure \ref{fig:4order}. We can see that the detection bounding box can be more compact, and thus the textual feature can be more accurately cropped for subsequent recognition.
\end{itemize}

\begin{figure}[!t]
  \centering
  \centerline{\includegraphics[width=0.5\textwidth
  ]
  {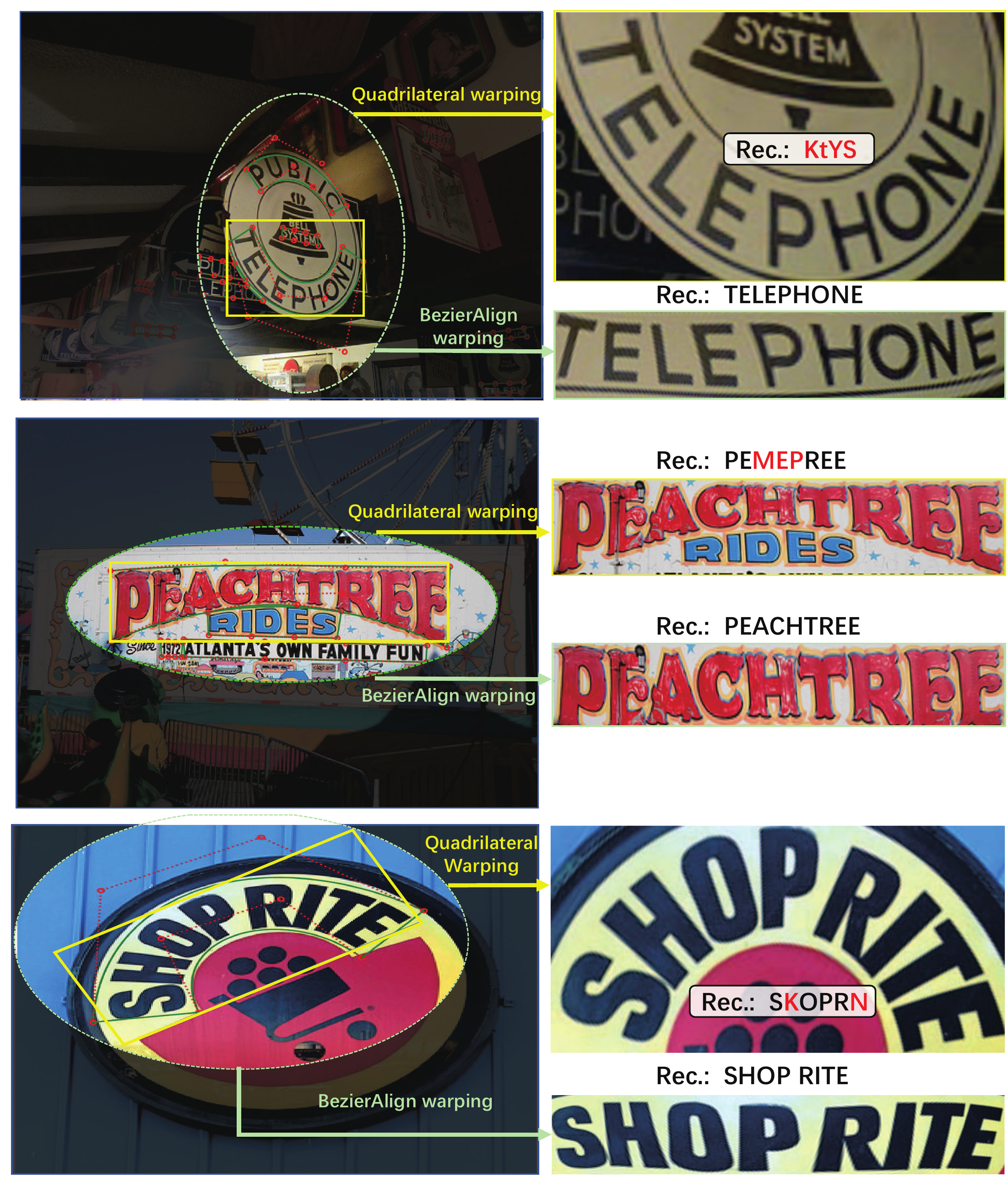}}
  \caption{\textbf{Qualitative recognition results of the quadrilateral sampling method and BezierAlign.} Left: original image.
  Top right:
  results by using quadrilateral sampling.
  Bottom right:
  results by using BezierAlign. }\label{fig:align_vis}
\end{figure}

We further evaluate BezierAlign by comparing it with previous sampling methods, shown in Figure \ref{fig:Bezier_align}. For fair and fast comparison, we use a small training and testing scale. The results shown in Table \ref{tab:exp_BezierAlign} demonstrate that the BezierAlign can dramatically improve the end-to-end results. Qualitative examples are shown in Figure \ref{fig:align_vis}. Another ablation study is conducted to evaluate the time consumption of Bezier curve detection, and we observe that the Bezier curve detection only introduces negligible computation overhead compared with standard bounding box detection.

\begin{table}[!t]
  \centering
  \newcommand{\tabincell}[2]{\begin{tabular}{@{}#1@{}}#2\end{tabular}}
  \caption{Ablation study for BezierAlign. Horizontal sampling follows \cite{li2017towards}, and quadrilateral sampling follows \cite{he2018end}.}
  \label{tab:exp_BezierAlign}
  \small
  \begin{tabular}{c| l |c}
     \hline
     Methods & Sampling method & F-measure (\%)  \\
     \hline
     baseline & +  Horizontal Sampling  & 38.4 \\
     baseline & +  Quadrilateral Sampling  & 44.7 \\
     baseline & +  BezierAlign  & 61.9 \\
     \hline
  \end{tabular}
\end{table}

\begin{table*}[!t]
  \caption{Ablation study on both Total-Text and SCUT-CTW1500 datasets.  Attn: attention recognition model. 4O: using 4th order Bezier Curve.}
  \label{tab:ablative}
  \centering
  \newcommand{\tabincell}[2]{\begin{tabular}{@{}#1@{}}#2\end{tabular}}
  \small
  \begin{tabular}{c|ccccc|c|c|c|c|c}
    \hline
    \multirow{2}*{ }  & \multicolumn{5}{c|}{Components} & \multicolumn{3}{c|}{Total-Text} &  \multicolumn{2}{c}{SCUT-CTW1500}\\
        \cline{2-11}
        & Attn & biFPN & CoordConv & AET & 4O & E2E Hmean & Impr. & FPS & E2E Hmean & Impr.\\
    \hline
    baseline (ctc) \cite{liu2020abcnet} &  &  &  &  & & 63.5 & - & 13 & 45.0 & -  \\
    \hline
    baseline+  & $\checkmark$ &  &  &  & & 66.2 & \bf  $\uparrow$ 2.7\%  & 11 & 52.9 & \bf  $\uparrow$ 7.9\%   \\
    baseline+  & $\checkmark$ & $\checkmark$ &  &  & & 68.1 & \bf   $\uparrow$ 4.6\% & 10 & 54.5 & \bf   $\uparrow$ 9.5\% \\
    baseline+  & $\checkmark$ & & $\checkmark$ &  &  & 69.0 & \bf   $\uparrow$ 5.5\% & 11 & 55.8 & \bf   $\uparrow$ 10.8\% \\
    baseline+  & $\checkmark$ &  & &  $\checkmark$ &  & 67.4 & \bf   $\uparrow$ 3.9\% & 11 & 54.6 & \bf   $\uparrow$ 9.6\% \\
    baseline+  & $\checkmark$ &  &  & & $\checkmark$ & 67.0 & \bf   $\uparrow$ 3.5\% & 11 & 54.4 & \bf   $\uparrow$ 9.4\% \\
    \hline
    ABCNet v2   & $\checkmark$ & $\checkmark$ & $\checkmark$ &  $\checkmark$ & & 70.4  & \bf   $\uparrow$ 6.9\% & 10  & 57.5 & \bf   $\uparrow$ 12.5\% \\
    \hline
  \end{tabular}
\end{table*}

\begin{table*}[!t]
  \caption{Detection results on Total-text, SCUT-CTW1500, MSRA-TD500, and ICDAR 2015 datasets.}
  \label{tab:cp_sota_det}
  \centering
  \newcommand{\tabincell}[2]{\begin{tabular}{@{}#1@{}}#2\end{tabular}}
\footnotesize
  \begin{tabular}{r |ccc|ccc|ccc|ccc|ccc}
    \hline
    \multirow{2}*{Methods}  & \multicolumn{3}{c|}{Total-Text} & \multicolumn{3}{c|}{SCUT-CTW1500} &  \multicolumn{3}{c|}{MSRA-TD500}&
    \multicolumn{3}{c|}{ICDAR 2015} & \multicolumn{3}{c}{ReCTS}\\
        \cline{2-16}
        & R  & P & H & R  & P & H & R  & P & H &R&P&H &R&P&H \\
    \hline
    Seglink \cite{shi2017detecting} &  30.3  &  23.8  & 26.7   &  48.4  & 38.3  & 42.8   & 70.0  & 86.0  & 77.0 &76.8&73.1&75.0  &-&-&-\\
    DMPNet \cite{liu2017deep} &  -  &  -  & -    & 61.7  & 63.9  & 62.7 &  -  & -   & - &68.2&73.2&70.6  &-&-&-\\
    CTD+TLOC \cite{liu2019curved} &  74.5 & 82.7  & 78.4  & 85.3  & 67.9  & 75.6  & 73.9  & 83.2  & 78.3   &77.1&84.5&80.6  &-&-&-\\
    TextSnake \cite{long2018textsnake}  & 74.5  & 8.7  & 78.4  & 77.8  & 82.7  &  80.1 & 81.7  & 84.2  & 82.9  &84.9&80.4&82.6  &-&-&-\\
    EAST \cite{zhou2017east} & 50.0 & 36.2 & 42.0 & 49.7 & 78.7 & 60.4 & 67.4 & 87.3 & 76.1  &78.3&83.3&80.7 &73.7&74.3&74.0\\
    He    \etal     \cite{he2018end}  &-&-&-  &-&-&-  &61.0&71.0&69.0  &-&-&-  &-&-&-\\
    DeepReg \cite{he2017deep} &-&-&-  &-&-&-  &70.0&77.0&74.0  &-&-&-  &-&-&-\\
    Textboxes++ \cite{liao2018textboxes++} & - & - & - & - & - & - & - & - & - &78.5&87.8&82.9  &-&-&-\\
    LSE \cite{tian2019learning}  & -  & -  & -  & 77.8  & 82.7  &  80.1 & 81.7  & 84.2  & 82.9  &-&-&- &-&-&-\\
    ATTR \cite{wang2019arbitrary}  & 76.2  &  80.9 & 78.5   &  80.2 & 80.1  & 80.1  &  82.1 & 85.2  &  83.6 &83.3&90.4&86.8  &-&-&-\\
    MSR \cite{xue2019msr}   &  73.0 &  85.2 & 78.6  & 79.0  & 84.1  &  81.5 & 76.7  & 87.4  & 81.7  &-&-&-  &-&-&-\\
    TextDragon \cite{feng2019textdragon}  & 75.7  & 85.6  & 80.3  & 82.8  & 84.5  & 83.6  &  - & -  & - &-&-&-  &-&-&-\\
    TextField \cite{xu2019textfield} & 79.9 & 81.2 & 80.6 & 79.8 & 83.0 & 81.4 & 75.9 & 87.4 & 81.3 &80.1&84.3&82.4  &-&-&-\\
    PSENet-1s \cite{wang2019shape}  & 78.0 & 84.0 & 80.9 & 79.7 & 84.8 & 82.2 & - & - & -  &85.5&88.7&87.1  &83.9&87.3&85.6\\
    Seglink++ \cite{tang2019seglink++}& 80.9 & 82.1 & 81.5 & 79.8 & 82.8 & 81.3 & - & - & - &80.3&83.7&82.0  &-&-&-\\
    LOMO \cite{zhang2019look}  & 79.3 & 87.6 & 83.3 & 76.5 & 85.7 & 80.8 & - & - & -  &83.5& \bf{91.3}&87.2 &-&-&-\\
    CRAFT \cite{baek2019character} & 79.9 & 87.6 & 83.6 & 81.1 & 86.0 & 83.5 & 78.2 & 88.2 & 82.9 &84.3&89.8&86.9  &-&-&-\\
    PAN \cite{wang2019efficient} & 81.0 & 89.3 & 85.0 & 81.2 & 86.4 & 83.7 &  \bf{83.8} & 84.4 & 84.1 &81.9&84.0&82.9  &-&-&-\\
    Mask TTD \cite{liu2019arbitrarily} & 74.5 & 79.1 & 76.7 & 79.0 & 79.7 & 79.4 & 81.1 & 85.7 & 83.3  & \bf{87.6}&86.6&87.1 &-&-&-\\
    CounterNet \cite{wang2020contournet} & 83.9 & 86.9 & 85.4 &  \bf{84.1} & 83.7 & 83.9 & - & - & - &86.1&87.6&86.9 &-&-&-\\
    DB \cite{liao2020real} &82.5&87.1&84.7  &80.2&86.9&83.4  &77.7&76.6&81.9 &82.7&88.2&85.4 &-&-&-\\
    Mask TextSpotter \cite{liao2019mask} &82.4&88.3&85.2 &-&-&- &-&-&- &87.3&86.6&87.0 & \bf{88.8}&89.3&89.0\\
    DRRN \cite{zhang2020deep}  & \bf{84.9} & 86.5 & 85.7 & 83.0 &  \bf{85.9} & 84.5 & 82.3 & 88.1 & 85.1 &84.7&88.5&86.6  &-&-&-\\
    \hline
    ABCNet v2 & 84.1 &\bf{90.2} & \bf{87.0} & 83.8 & 85.6 & \bf{84.7} &81.3& \bf{89.4}&\bf{85.2} &86.0&90.4&\bf{88.1} &87.5& \bf{93.6}&\bf{90.4}\\
    \hline
  \end{tabular}
\end{table*}

\begin{table*}[!t]
  \caption{End-to-End text spotting results on Total-Text, and SCUT-CTW1500. * represents the results are from \cite{feng2019textdragon}. ``None'' represents lexicon-free. ``Strong Full'' represents that we  use all the words appeared in the test set. ``S'', ``W'', and ``G'' represent recognition with ``Strong'', ``Weak'', and ``Generic'' lexicon, respectively. FPS is for reference only, as it %
  can
  be varied from different settings and devices. Here * represents multi-scale results.
  }
  \label{tab:cp_sota_e2e}
  \centering
  \newcommand{\tabincell}[2]{\begin{tabular}{@{}#1@{}}#2\end{tabular}}
\small
  \begin{tabular}{ r |cc|cc|ccc|c|c}
    \hline
    \multirow{2}*{Methods}  & \multicolumn{2}{c|}{Total-Text} & \multicolumn{2}{c|}{SCUT-CTW1500} &
    \multicolumn{3}{c|}{ICDAR 2015 End-to-End}  &
    \multicolumn{1}{c|}{ReCTS}  & FPS\\
        \cline{2-10}
        & None & Full & None & Full  & S & W & G &1-NED \\
    \hline
    TextBoxes++ \cite{liao2018textboxes++} &36.3&48.9 &-&-  &73.3&65.9&51.9 &-  &1.4\\
    Mask TextSpotter'18 \cite{lyu2018mask}  &52.9&71.8 &-&-  &79.3&73.0&62.4 &-  &2.6\\
    TextNet \cite{sun2018textnet}  &54.0&- &-&-  &-&-&- &-  &2.7\\
    Li    \etal     \cite{li2019towards}  &57.8&- &-&-  &-&-&- &-  &1.4\\
    Deep Text Spotter \cite{busta2017deep} &-&- &-&-  &54.0&51.0&47.0  &-  &-\\
    Mask TextSpotter'19 \cite{liao2019mask}  &65.3&77.4  &-&- &83.0&77.7&73.5 &\bf{67.8}  &2.0\\
    Qin    \etal     \cite{qin2019towards}  &67.8&- &-&-  &-&- &-  &- &4.8\\
    CharNet \cite{xing2019convolutional}  &-&-  &-&- &80.1&74.5&62.2 &-  &1.2\\
    FOTS \cite{liu2018fots} &-&- &21.1&39.7  &83.6&79.1&65.3 &50.8  &-\\
    RoIRotate* \cite{feng2019textdragon} &-&- &38.6&70.9  &82.5&79.2&65.4 &- &-\\
    TextDragon \cite{feng2019textdragon} &48.8&74.8 &39.7&72.4  &82.5&78.3&65.2 &- &2.6\\
    ABCNet \cite{liu2020abcnet} &64.2&75.7 &45.2&74.1  &-&-&- &- &\bf{17.9}\\
    Boundary TextSpotter \cite{boundary2020} &-&- &-&-  &79.7&75.2&64.1 &- &-\\
    Craft \cite{baek2020character} &\bf{78.7}&- &-&-  &83.1&82.1&74.9 &- &5.4\\
    Mask TextSpotter v3 \cite{liao2020masktext} &71.2&78.4 &-&-  &83.3&78.1&74.2 &- &2.5\\
    Feng    \etal     \cite{feng2020residual} &55.8&79.2 &42.2&74.9 &\bf{87.3}&\bf{83.1}&69.5 &-&7.2\\
    \hline
    ABCNet v2  & 70.4 & 78.1  & 57.5 & 77.2  & 82.7 & 78.5 & 73.0  & 62.7 & 10\\
    ABCNet v2*  & 73.5 & \bf{80.7} &\bf{58.4}& \bf{79.0}  &83.0&80.7&{\bf{75.0}}  &65.4 & -\\
    \hline
  \end{tabular}
\end{table*}

\subsection{Comparison with State-of-the-art}
We compare our method to previous methods on both detection and end-to-end text spotting tasks. An optimal setting including inference thresholds and testing scale is decided using grid search. For the detection task, we conduct experiments on four datasets including two arbitrarily-shaped datasets (Total-Text and SCUT-CTW1500), two multi-oriented datasets (MSRA-TD500, and ICDAR 2015) and one bilingual dataset ReCTS. The results in Table \ref{tab:cp_sota_det} demonstrate that our method can achieve state-of-the-art performance on all four datasets, outperforming previous state-of-the-art methods.

For end-to-end scene text spotting tasks, \textit{ABCNet v2 achieves the best performance on SCUT-CTW1500 and ICDAR 2015 datasets, significantly outperforming previous methods. }
The results are shown in Table \ref{tab:cp_sota_e2e}. Although our method is worse than Mask TextSpotter \cite{liao2019mask} in terms of 1-NED on the ReCTS dataset, we argue that we do not use the provided character-level bounding box and ours shows clear advantages in terms of the inference speed. On the other hand, ABCNet v2 can still achieve
better detection performance compared to Mask TextSpotter \cite{liao2019mask} according to Table \ref{tab:cp_sota_det}.

Qualitative results of the test sets are shown in Figure \ref{fig:FIRES}. From the figure, we can see that ABCNet v2 achieves a powerful recall ability for various text including horizontal, multi-oriented, and curved text, or long and dense text presentation styles.

\subsection{Comprehensive Comparison with Mask TextSpotter v3}
\begin{figure}[!t]
   \centering
   \centerline{\includegraphics[width=0.5\textwidth
   ]{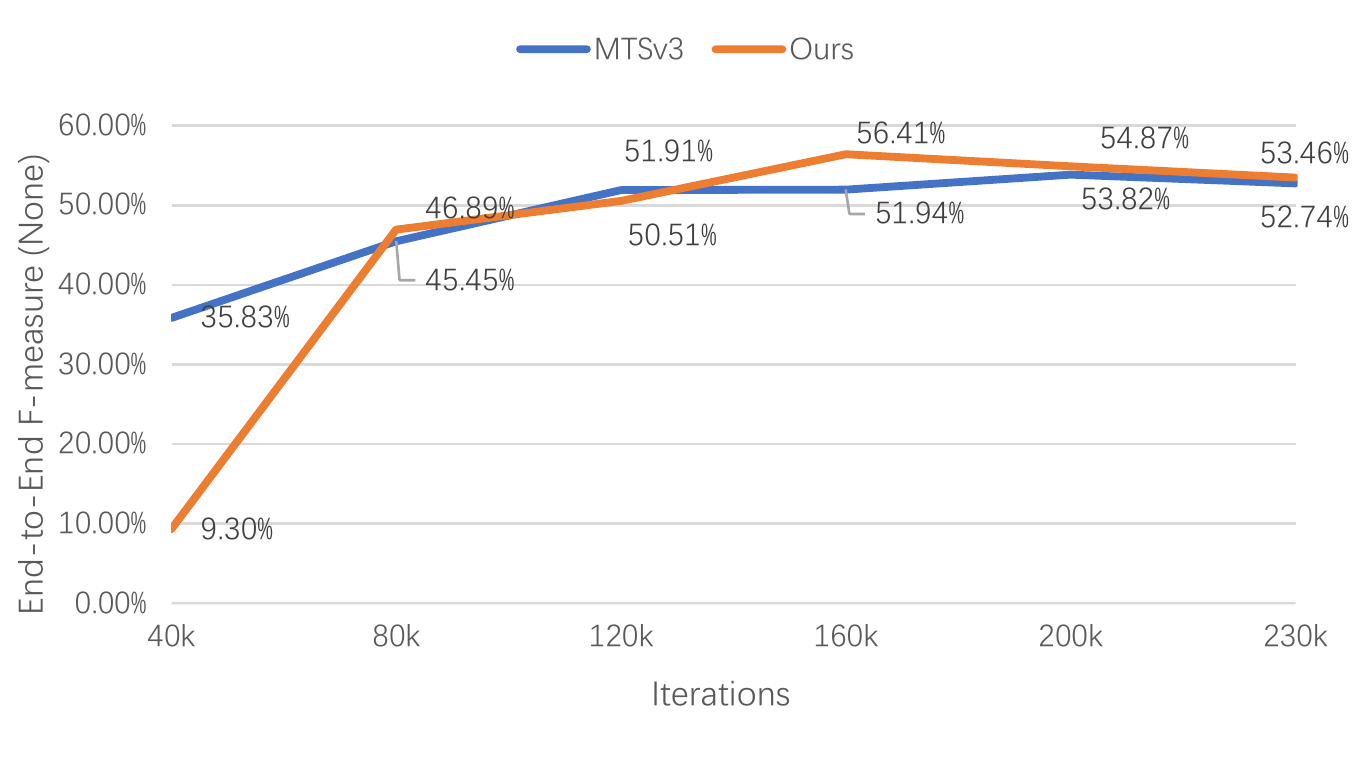}}
   \caption{
   \textbf{Comparison with Mask TextSpotter v3 using only the training set of TotalText.
   MTSv3: Mask TextSpotter v3 \cite{liao2020masktext}.}
   }\label{fig:acc}
\end{figure}

{\bf Comparison using few data.} We find out the proposed method can achieve a promising spotting result using only a small amount of training data. To validate the effectiveness, we use the official code of the Mask TextSpotter v3 \cite{liao2020masktext}, and conduct experiments following the same setting by training the model with only the official training data of TotalText. Specifically, the optimizer and the learning rate (0.002) of our method are set to the same as that of Mask TextSpotter v3. The batch sizes are set to 4, and both methods are trained with 230K iterations. To ensure the best setting of the both methods, Mask TextSpotter v3 is trained by minimum sizes of 800, 1000, 1200, and 1400, and maximum size of 2333. Testing is conducted with a minimum size of 1000 and maximum size of 4000. Our method is trained by a minimum size from 640 to 896 with the interval of 32, and the maximum size is set to 1600. To stabilize the training, AET strategy is not used for the few-shot training.
Testing is conducted with a minimum size of 1000, and maximum size of 1824. We also conduct grid search to find the best threshold for the Mask TextSpotter v3. The results of different iterations are shown in Figure \ref{fig:acc}. We can see that although Mask TextSpotter v3 converges faster at the beginning, the final result of our method is better (56.41\% vs.\  53.82\%).

{\bf Comparison using large-scale data.} We also use sufficient training data for a more thorough comparison with Mask TextSpotter v3. Formally, we carefully train Mask TextSpotter v3 using the Bezier Curve Synthetic Dataset (150k), MLT (7k), and TotalText, which are exact the same as our method. The training scales, batch sizes, the number of iterations, and others are all set to the same as mentioned in the Section \ref{subsec:Imple_details}. Grid search is also used to find the best threshold for the Mask TextSpotter v3. The results are shown in
Table~\ref{tab:exp_mtsv3_large}, from which we can see that Mask TextSpotter v3 outperforms ABCNet v1 by 0.9 in terms of F-measure, and ABCNet v2 can outperform Mask TextSpotter v3 (65.1\% vs.\  70.4\%). The inference time is measured under the same testing scale (Maximum size is set to 1824) and device (RTX A40), which further demonstrates the effectiveness of our method.

\begin{table}[!t]
  \centering
  \newcommand{\tabincell}[2]{\begin{tabular}{@{}#1@{}}#2\end{tabular}}
  \caption{Comparison with Mask TextSpotter v3 using large-scale training set. MTSv3: Mask TextSpotter v3 \cite{liao2020masktext}.}
  \label{tab:exp_mtsv3_large}
  \small
  \begin{tabular}{c|c|c|c}
     \hline
      & MTSv3 & ABCNet v1 & ABCNet v2  \\
     \hline
     F-measure (\%) & 65.1 & 64.2 & {\bf 70.4}  \\
     \hline
     FPS & 2.5 & {\bf 14.3} & 8.7 \\
     \hline
  \end{tabular}
\end{table}

\begin{table*}[htb!]
  \centering
  \small
	\caption{Quantization results of ABCNet v2 (ResNet-18 as the backbone). ``A/W'' indicates the bit width configuration of the activation and weight, respectively. \textsuperscript{\dag} implies the case is trained with progressive training strategy. FPS is based on profiling data on single Nvidia 2080TI GPU.}
	\label{tab:quantization-resnet18}
		\begin{tabular}{ c | c | c c c | c c c |c}
			\hline
			\multirow{2}{*}{Data set} & \multirow{2}{*}{A/W} &
			\multicolumn{3}{c|}{End-to-end Results} &
			\multicolumn{3}{c|}{Detection only Results} &
			\multirow{2}{*}{FPS} \\
			& & precision (\%)& recall (\%)& hmean (\%) & precision (\%)& recall (\%)& hmean(\%) & \\
            \hline
            Pretrain for Total-Text  & 32/32 & 66.1 & 51.0 & 57.6 & 83.3 & 76.7 & 79.9 & 16.4 \\
            Finetune on Total-Text& 32/32 & 70.3 & 64.4 & 67.2 & 86.2 & 82.9 & 84.5 & 16.4 \\
            Finetune on CTW1500   & 32/32 & 58.8 & 48.3 & 53.0 & 88.2 & 81.4 & 84.7 & 37.9 \\
            \hdashline
            Pretrain for Total-Text  & 4/4 & 67.2 & 52.4 & 58.9 & 83.4 & 77.7 & 80.5 & 25.9\\
            Finetune on Total-Text& 4/4 & 70.7 & 62.8 & 66.5 & 86.0 & 82.3 & 84.1 & 25.9\\
            Finetune on CTW1500   & 4/4 & 58.0 & 47.3 & 52.2 & 87.2 & 81.0 & 84.0 & 47.8\\
            \hdashline
            Pretrain for Total-Text & 1/1 & 62.0 & 34.4 & 44.3 & 82.0 & 63.6 & 71.6 & 29.5\\
            Pretrain for Total-Text & 1/1\textsuperscript{\dag} & 65.6 &48.5 & 55.8 & 81.8 & 74.9 & 78.2 & 29.5 \\
            Finetune on Total-Text & 1/1\textsuperscript{\dag} & 71.2 & 60.8 & 65.5 & 88.0 & 82.2 & 85.0 & 29.5 \\
            Finetune on CTW1500 & 1/1\textsuperscript{\dag} & 58.2 & 46.5 & 51.7 & 85.1 & 80.4 & 82.7 & 51.2 \\
            \hline
	\end{tabular}
\end{table*}

\subsection{Limitations}
We further conduct error analysis on the incorrectly predicted samples. We observe two types of common errors that may
limit the scene text spotting performance of ABCNet v2.

\begin{figure}[!h]
  \centering
  \centerline{\includegraphics[
  width=0.4\textwidth
  ]{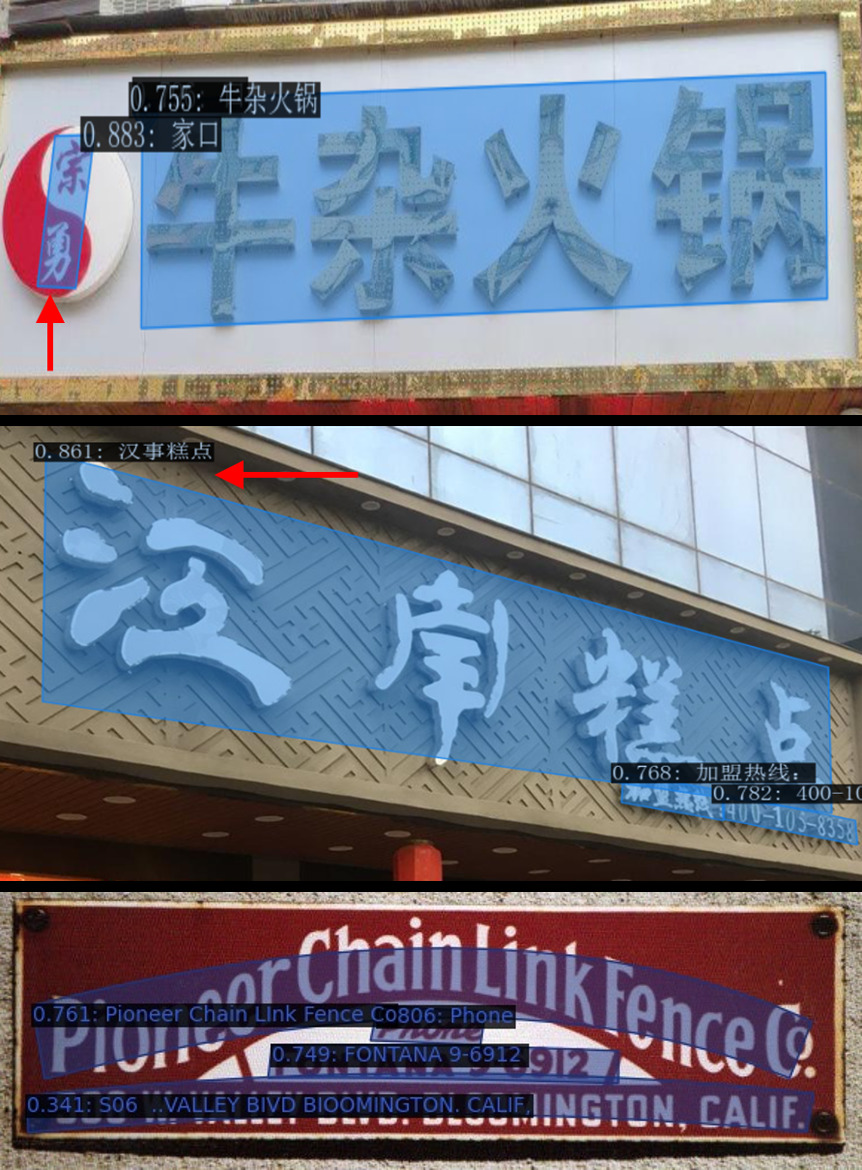}}
  \caption{\textbf{Error analysis of ABCNet v2.}}\label{fig:error}
\end{figure}

The first one is shown in the example of Figure \ref{fig:error}. The text instance contains two characters. For each character, the reading order is from left to right. But for the whole instance, the reading order is from up to bottom. As the Bezier curve is interpolated in the longer side of the text instance, the BezierAlign feature would be a rotated  feature compared to its original feature, which can result in a completely different meaning. On the other hand, such cases only consist a minority of the whole training set, which is prone to be mistakenly recognized or predicted as an unseen category, as represented by ``$\square$'' for the second character.

The second errors happen in different fonts, as shown in the middle of Figure \ref{fig:error}. The first two characters are written in unusual calligraphy fonts, making it difficult to recognize.
In general, this challenge can only be alleviated with more training images.

We also find out that there is an extremely curved case in the test set of the CTW1500, where more than three crests exist in the same text instance, as shown in the third row of the Figure \ref{fig:error}. In such a case, the lower order such as cubic Bezier curve may be limited, as the character ``i'' is incorrectly recognized as the uppercase ``I'' because of the inaccurate shape representation. However, such cases are rarely seen, especially for those datasets using word-level bounding box.

\subsection{Inference Speed}
To further test the potential real-time performance of the proposed method, we exploit quantization techniques to significantly improve the inference speed. The baseline model adopts the same setting as the baseline with an attention based recognition branch shown in Table \ref{tab:ablative}. The backbone is replaced by ResNet-18, which significantly improves the speed with only marginal accuracy reduction.

The performance of the quantized network with various quantization bit configurations ($4/1$-bit) are reported in Table \ref{tab:quantization-resnet18}. Full precision performance is also listed for comparison. To train the quantized network, we first pretrained the low-bit model on the synthetic dataset. Then, we fine-tune the network on dedicated dataset $\rm TotalText$ and $\rm CTW1500$ for better performance.

Results of accuracy for both the pretrained model and fine-tuned model are reported. During the pre-training, $260$K iterations are trained with batch size to be 8. Initial learning rate is set to be 0.01 and divided by 10 at $160$K and $220$K iteration. For the fine-tuning on $\rm TotalText$ dataset, batch size remains 8 and the initial learning rate is set to be 0.001. Only $5$K iterations are fine-tuned. The batch size and initial learning rate are the same when fine-tuning on $\rm CTW1500$ dataset. However, the number of total iterations is set to be $120k$ and the learning rate is divided by 10 at iteration $80k$. Similar with previous quantization work, we quantize all the convolutional layer in the network, except input and output layers. If not specifically stated, networks are initialized with the full precision counterpart.

From Table \ref{tab:quantization-resnet18}, \textit{we can learn that the 4-bit models by our quantization method are able to obtain comparable performance with the full precision counterparts.}
For example, end-to-end $\rm hmean$ of 4-bit model pretrained on synthetic dataset is even better than that of the full precision model (57.6\% vs.\  58.9\%). After fine-tuning, end-to-end $\rm hmean$ of 4-bit model on $\rm TotalText$ and $\rm CTW1500$ is only 0.7\% (67.2\% vs.\  66.5\%) and 0.8\% (53.0\% vs.\  52.2\%) lower than that of the full precision model, respectively.

The fact that almost no performance drops for 4-bit models (the same observation is also made on image classification and objection detection tasks \cite{Li_2019_CVPR}) indicates the considerable redundancy in the full precision scene text spotting model.

However, the performance has a sizable drop for the binary network, with the end-to-end $\rm hmean$ to be only 44.3\%. To compensate, we propose to train the BNN (binary neural network) model with progressive training, in which the quantization bit width is progressively decreased (\textit{e.g.}, 4-bit $\to$ 2bit $\to$ 1-bit).
With the new training strategy (the ones with \textsuperscript{\dag} in the table), the performance of the binary network is significantly improved. For example, the end-to-end $\rm hmean$ trained on synthetic dataset by BNN model is only 1.8\% (57.6\% vs.\  55.8\%) lower than the full precision counterpart.

Apart from the performance evaluation, we also compare the overall speed of the quantized model against full-precision models. In practice, only the quantized convolution layers are accelerated with other layers, such as $\rm LSTM$ keeping in full precision. It can be learned from Table \ref{tab:quantization-resnet18} that, with limited performance drop, the binary network of ABCNet v2 is able to run in real-time for both $\rm TotalText$ and $\rm CTW1500$ datasets.

\section{Conclusion}
\label{sec:conclude}
We have proposed ABCNet v2---a real-time end-to-end method that uses Bezier curves for arbitrarily-shaped scene text spotting. By reformulating arbitrarily-shaped scene text using parameterized Bezier curves, ABCNet v2 can detect arbitrarily-shaped scene text with Bezier curves.
Our method introduces negligible computation cost compared with standard bounding box detection. With such Bezier curve bounding boxes, we can naturally connect a light-weight recognition branch via a new BezierAlign layer, which is critical for accurate feature extraction, especially for curved text instances.

Comprehensive experiments on various datasets demonstrate the effectiveness of the proposed components, including using the attention recognition module, biFPN structure, coordinate convolution, and a new adaptive end-to-end training strategy. Finally, we propose to apply quantization techniques for deploying our model for real-time tasks,
showing the great potential for a wide range of applications.

{
\bibliographystyle{ieeetr}
 \bibliography{CSRef}

}

\vfill
\end{document}